%% file: main.tex
\documentclass[10pt,twocolumn,letterpaper]{article}

\usepackage[pagenumbers]{wacv} 

\input{preamble}

\definecolor{wacvblue}{rgb}{0.21,0.49,0.74}
\usepackage[pagebackref,breaklinks,colorlinks,allcolors=wacvblue]{hyperref}
\usepackage{graphicx}
\usepackage{array}
\usepackage{paralist} 
\usepackage[skip=3pt]{caption}
\usepackage{caption}
\usepackage{enumitem}   

\def\wacvPaperID{1023} 
\def\confName{WACV}
\def\confYear{2026}

\title{MapleGrasp: Mask-guided Feature Pooling for Language-driven \\ Efficient Robotic Grasping}

\author{Vineet Bhat \and Naman Patel \and Prashanth Krishnamurthy \and Ramesh Karri \and Farshad Khorrami \\
\\ New York University Tandon School of Engineering\\
Brooklyn, NY, USA\\
{\tt\small vrb9107@nyu.edu}
}
\begin{document}
\maketitle
\input{sec/0_abstract}    
\input{sec/1_intro}
\input{sec/2_relatedworks}

\input{sec/3_maskguidedpooling}

\input{sec/4_refgraspnetdataset}

\input{sec/5_experiments}
\input{sec/6_conclusion}
{
    \small
    \bibliographystyle{ieeenat_fullname}
    \bibliography{main}
}

\end{document}

%% file: preamble.tex
\usepackage{pifont}
\usepackage{booktabs}
\newcommand{\cmark}{\ding{51}} 
\newcommand{\xmark}{\ding{55}} 


%% file: sec/0_abstract.tex



\begin{abstract}
Robotic manipulation of unseen objects via natural language commands remains challenging. Language driven robotic grasping (LDRG) predicts stable grasp poses from natural language queries and RGB-D images. We propose MapleGrasp, a novel framework that leverages mask-guided feature pooling for efficient vision-language driven grasping. Our two-stage training first predicts segmentation masks from CLIP-based vision-language features. The second stage pools features within these masks to generate pixel-level grasp predictions, improving efficiency, and reducing computation. Incorporating mask pooling results in a 7$\%$ improvement over prior approaches on the OCID-VLG benchmark. Furthermore, we introduce RefGraspNet, an open-source dataset eight times larger than existing alternatives, significantly enhancing model generalization for open-vocabulary grasping. MapleGrasp scores a strong grasping accuracy of 89\% when compared with competing methods in the RefGraspNet benchmark. Our method achieves comparable performance to larger Vision-Language-Action models on the LIBERO benchmark, and shows significantly better generalization to unseen tasks. Real-world experiments on a Franka arm demonstrate 73$\%$ success rate with unseen objects, surpassing competitive baselines by 11$\%$. Code is available here: \url{https://github.com/vineet2104/MapleGrasp}
\end{abstract}



%% file: sec/1_intro.tex
\begin{figure}[!ht]
  \centering
   \includegraphics[width=\linewidth]{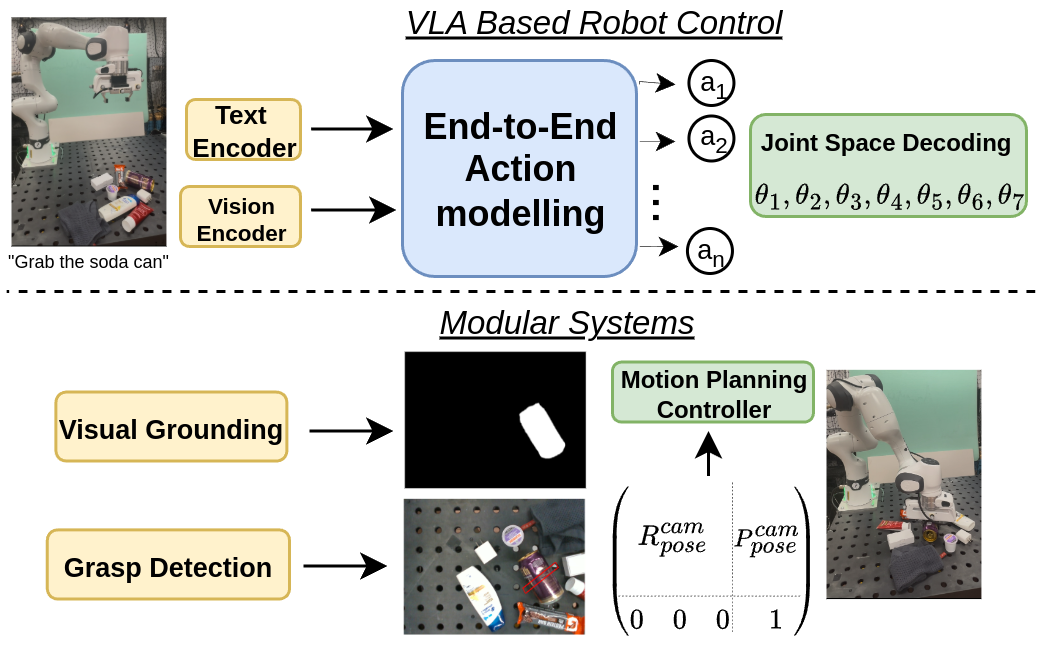}
   \caption{Modular systems generate object masks and grasp poses, utilizing a control algorithm to estimate joint space trajectories, while VLAs directly map vision and textual modalities to actions.}
   
   \label{fig:ldgd-approaches}
\end{figure}

\vspace{-15pt}
\section{Introduction}
\label{sec:intro}

The ability to perceive and interact with objects in an environment is a defining characteristic of human intelligence. As humans mature, we develop capabilities to manipulate novel objects, generalizing our understanding beyond previous experiences. Replicating this intelligence in robotics remains a formidable challenge. Robots often encounter unfamiliar objects and must adapt to new scenarios without prior knowledge~\cite{10711845}. Foundational models, particularly Vision-Language Models (VLMs), aim to bridge this gap by leveraging extensive pre-trained multi-modal representations to enhance robotic perception and interaction~\cite{li2024visionlanguage,zhao2024vlmpc}.

An autonomous robotic manipulation framework typically comprises four key stages: task planning, visual grounding, object manipulation, and goal-oriented placement. Recent advances have demonstrated the effectiveness of VLMs across these stages, incorporating large-scale pre-trained representations to improve generalization~\cite{10161041}. An alternative paradigm involves Vision-Language-Action (VLA) models, which directly predict trajectories for grasping and manipulating objects, bypassing the modular pipeline~\cite{Gao2023PhysicallyGV,rt22023arxiv}. We highlight the difference in these two frameworks in \cref{fig:ldgd-approaches}. While VLAs have shown remarkable performance in tasks across both simulation and real-world settings, their deployment in novel environments presents three key challenges: \textbf{i) Task-Specific Data Collection}: Collecting teleoperated or human demonstrations is often difficult and time-consuming for novel interactions~\cite{10341577}; \textbf{ ii) Expensive fine-tuning}: Large VLAs require task-specific fine-tuning, which is computationally expensive requiring hundreds of GPU hours. Although parameter-efficient fine-tuning (PEFT) techniques alleviate some costs, they restrict adaptability by modifying only a small subset of parameters~\cite{kim25finetuning}; \textbf{ iii) Limited generalization}: Fine-tuned VLAs often fail to generalize in dynamic environments, particularly when objects are repositioned or when visually similar distractor objects are introduced~\cite{ocidvlg}. Given these challenges, modular frameworks offer an attractive alternative that balances generalizability with ease of deployment. By decoupling perception, planning, and manipulation, modular architectures allow state-of-the-art models in each component to be independently improved and integrated as research advances. Furthermore, LLM-agentic frameworks can serve as high-level decision-makers, dynamically invoked perception and manipulation modules for task execution in novel environments~\cite{momallm24,rana2023sayplan,song2023llmplanner}. Compounded errors within modular frameworks can be mitigated using closed-loop feedback and error planning~\cite{curtis2024trust,pmlr-v205-huang23c}. 

In this work, we focus on a critical capability within modular frameworks, language-driven grasp detection. This task predicts a stable grasp pose for a specified object based on RGB-D images and a free-form language expression. An accurate LDRG model can enable low-level execution for high-level LLM planners, advancing fully autonomous agentic frameworks for human-robot interaction. Our proposed method, MapleGrasp, can predict both 4 DoF top-down and oriented 6 DoF grasps. Our contributions include:
\begin{compactenum}
    \item A novel two-stage training architecture for LDRG: initial text-referred object mask prediction, followed by mask-guided feature pooling for grasp refinement.
    \item Empirical evidence that restricting grasp predictions to mask-pooled regions leads to faster and more efficient training, while achieving better accuracy against previous methods by 7$\%$ on the OCID-VLG benchmark.
    \item Introduce RefGraspNet, a large-scale open-source benchmark comprising over 200 million grasps, eight times larger than existing datasets (\cref{tab:dataset_comparison}), consisting of challenging real-world scenarios for grasping. 
    \item Extensive comparisons with VLAs in a physics simulated environment demonstrate strong performance and improved generalization to novel tasks. MapleGrasp robustly transfers to real-world trials with the Franka arm, achieving 87$\%$ and 73$\%$ grasping success rates in seen and unseen environments, respectively, surpassing previous methods by more than 11$\%$.
\end{compactenum}


%% file: sec/2_relatedworks.tex
\section{Related Works}
\label{sec:related_works}

\noindent \textbf{Grasp Detection.} Robotic grasping with parallel grippers has been extensively studied in both the 2D and 3D domains. The 4-DoF grasp representation defines a grasp using four parameters: $(x,y)$ as the center of the grasp, $w$ as the grasp width, and $\theta$ as the grasp axis angle relative to a top-down hand orientation~\cite{detseg,jacquard,5980145}. The 6-DoF representation extends this to 3D, incorporating three degrees of translation and three degrees of rotation required for stable object grasping~\cite{9811961,acronym,9126187}. The use of RGB-D data has shown promising results for precision grasping, where CNN and ResNet-based architectures leverage visual features to predict grasp pose parameters~\cite{gr-convnet,graspanything,9341056,9691835}. 
Recent advancements in transformer architectures and self-supervised learning have bridged the gap between vision and textual modalities, shifting research towards incorporating language as an additional modality for object manipulation~\cite{10161041}. Language-driven grasp detection (LDGD) refers to the task of associating referring text expressions with RGB-D images to predict a suitable robotic grasp pose for object manipulation~\cite{graspanything++}. Two primary approaches have emerged for LDGD: i) Visual grounding of referring expressions: Predicts segmentation masks for the target object~\cite{deitke2024molmopixmoopenweights,UNINEXT,zhang2024evfsamearlyvisionlanguagefusion}, followed by pre-trained grasp detection networks to compute grasp poses~\cite{bhat2024hificsopenvocabularyvisual,liu2024ok,vlgrasp}. ii) End-to-end grasp pose estimation: Directly predicts grasp poses from input text and RGB-D images~\cite{YANG2024105280,10758335,NEURIPS2024_5367f6d5}.


\begin{figure*}[!h]
  \centering
   \includegraphics[width=\linewidth]{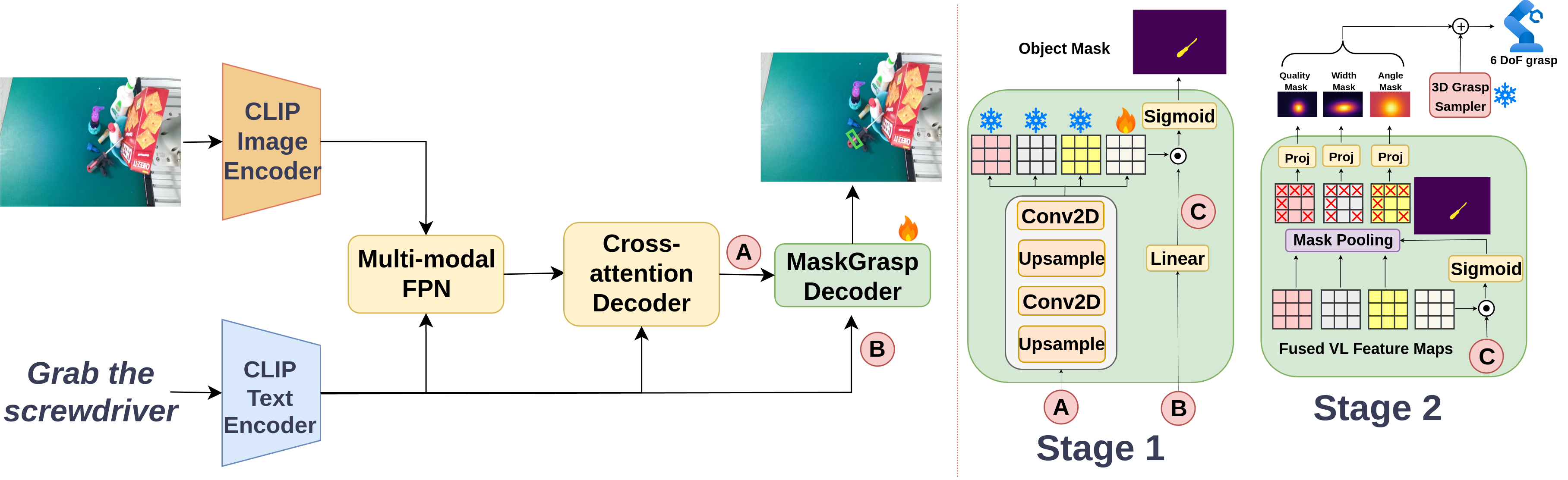}
   \caption{MapleGrasp: Our framework trains on fused vision-language embeddings for object segmentation, then applies mask pooling in grasp prediction heads. The predicted grasp maps can be used for both top-down and oriented grasping.}
   \label{fig:maplegrasp-arch}
\end{figure*}

\noindent \textbf{End-to-End Language-Driven Grasping.} Recent work has introduced diffusion models that learn a probabilistic diffusion process to generate grasp poses during inference~\cite{graspanything++,llgd,10.1007/978-3-031-72655-2_21}. Vision-language fusion provides an alternative approach, offering faster convergence and lower inference times~\cite{graspmamba,Li_2024_IROS,10161041}. The ETRG framework enhances vision-language feature fusion for grasp detection by integrating trainable adapters into frozen CLIP vision and text encoders~\cite{yu2024parameter}. MaskGrasp employs a tri-head architecture with cross-attention mechanisms using text embeddings, region-of-interest feature embeddings, and mask embeddings~\cite{maskgrasp}. GraspSAM adapts the Segment Anything Model~\cite{Kirillov_2023_ICCV} for grasp detection and utilizes GroundingDINO~\cite{liu2023grounding} to transform text expressions into object bounding boxes~\cite{noh2024graspsam}. GraspMolmo~\cite{deshpande2025graspmolmo} trains the Molmo VLM on the grasp point detection task followed by a grasp proposal sampler to identify accurate grasps for task oriented grasping. Our proposed architecture predicts grasp pose using quality, width and angle maps. Several studies have explored similar setups for LDGD~\cite{tziafas2024towards,ocidvlg,10891012}. These maps can be used to predict both 4 DoF and 6 DoF grasps by computing the grasp point on the object and overlapping with an independent grasp proposal network. 

In this work, we study mask pooling for grasp detection and demonstrate that localizing the grasp prior to predicting its angle and width leads to a more efficient learning process. This approach improves both training efficiency and test-time accuracy. Additionally, we present the first comprehensive comparison of LDRG approaches with large VLAs in the context of language-guided grasping tasks.

\noindent \textbf{Vision-Language-Action Models.} Direct modeling of vision and language modalities for joint-space robotic control has gained significant interest in recent research. Advances in vision-language models~\cite{Zhai_2023_ICCV,karamcheti2024prismatic,lu2024deepseekvl}, coupled with the availability of large-scale robotic datasets spanning diverse environments, hardware platforms, and tasks~\cite{open_x_embodiment_rt_x_2023}, have enabled the training of Vision-Language-Action (VLA) models for robotic manipulation~\cite{3dmvp,rvt2,li2024llara}. However, the large model sizes of VLAs often demand substantial GPU resources for fine-tuning and real-world deployment~\cite{10711245}. Recent research has sought to mitigate these computational challenges through parameter-efficient training techniques and action chunking~\cite{kim24openvla,kim25finetuning}, enabling deployment on smaller GPUs with real-time kernel compatibility.

%% file: sec/3_maskguidedpooling.tex
\vspace{-5pt}
\section{Methodology}
\label{sec:mask_guided_pixel_pooling}


We propose object-mask feature pooling to isolate relevant regions for grasp detection in clutter, detail our architecture (\cref{fig:maplegrasp-arch}), and outline an enhanced training objective.

\subsection{Language-Driven Grasp Detection}
\label{subsec:ldgd}
\begin{figure}[ht]
  \centering
  \setkeys{Gin}{width=0.45\linewidth}
  \begin{tabular}{cc}
    \includegraphics{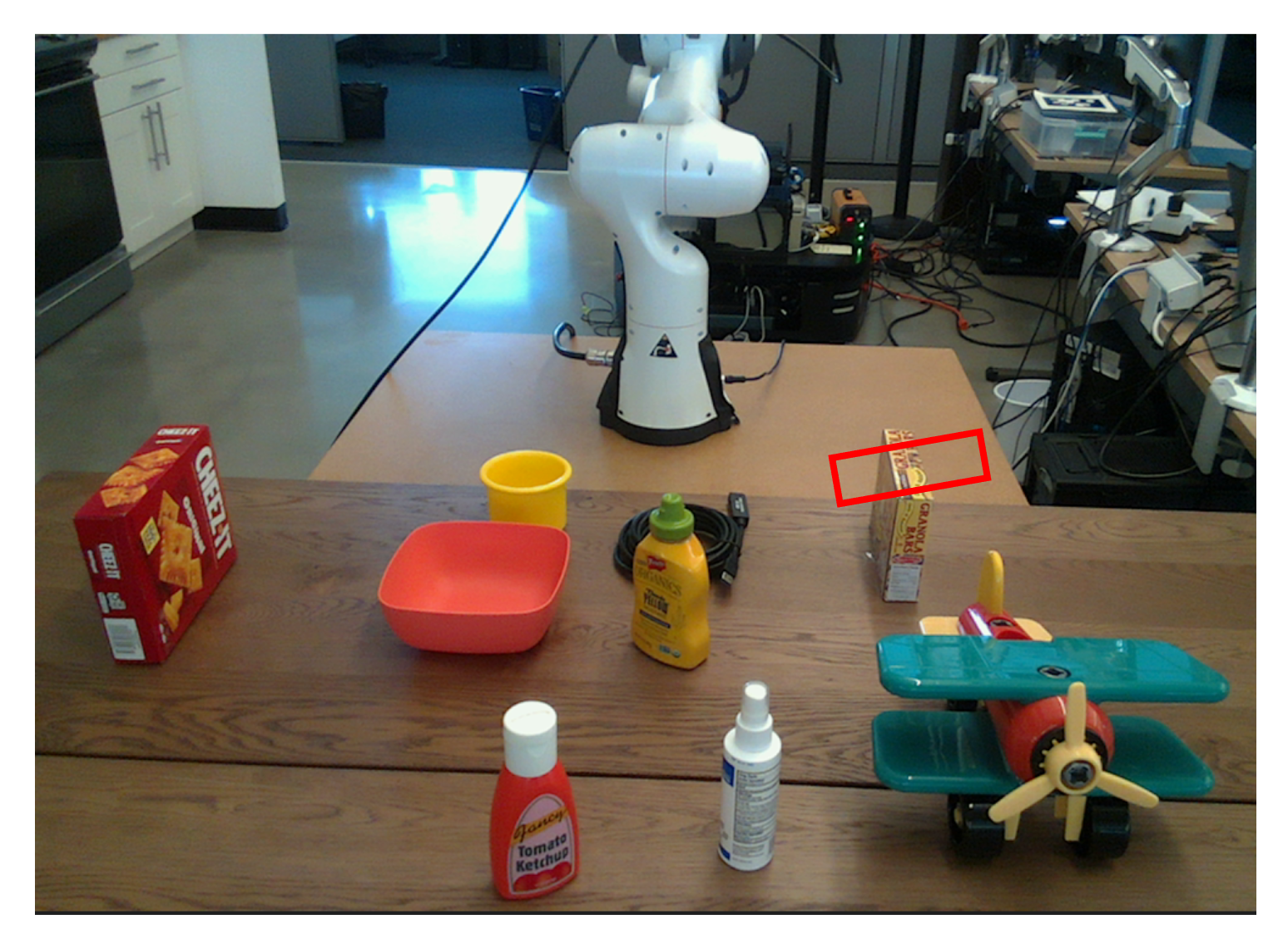} & 
    \includegraphics{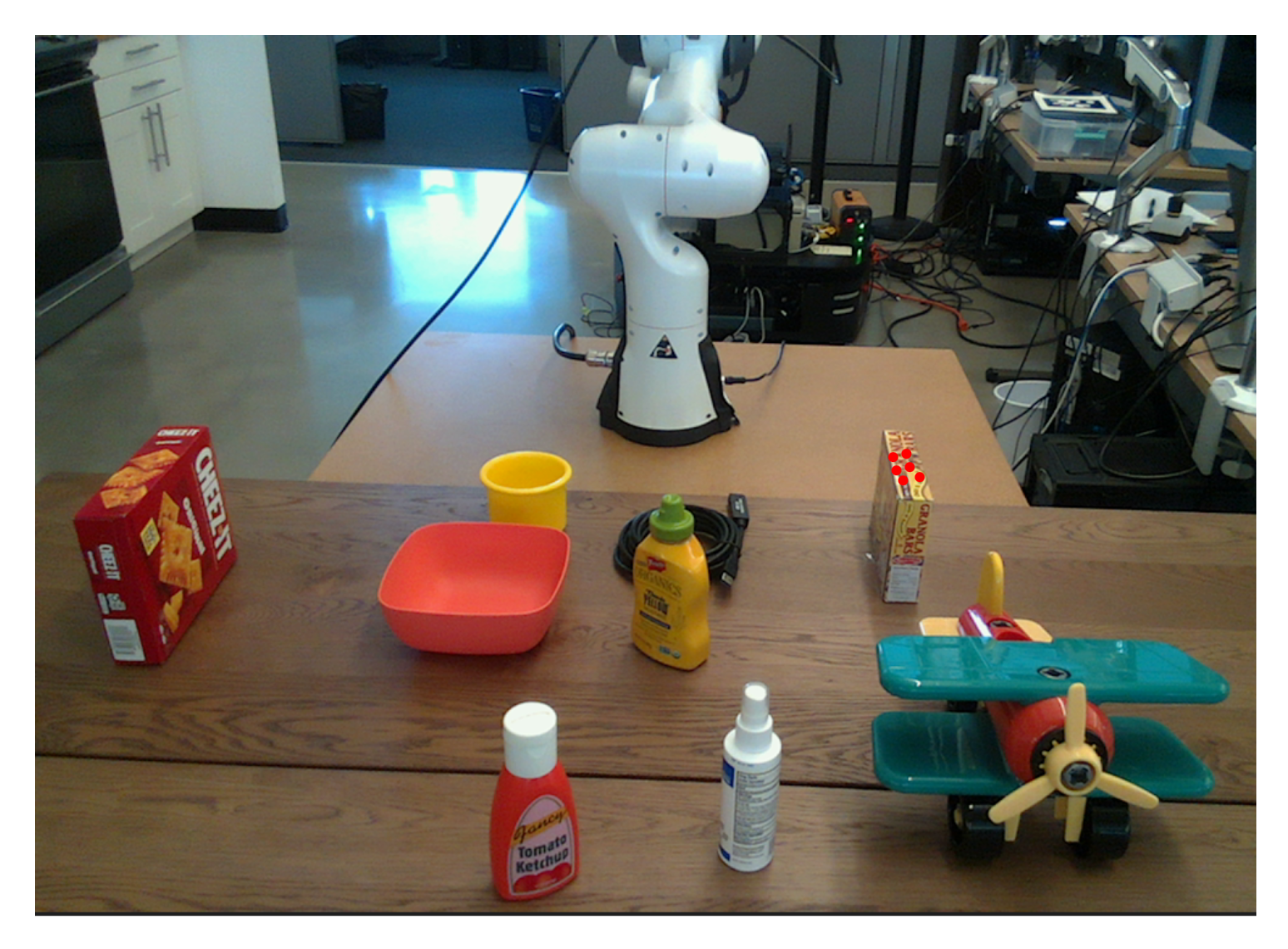} \\
    \includegraphics{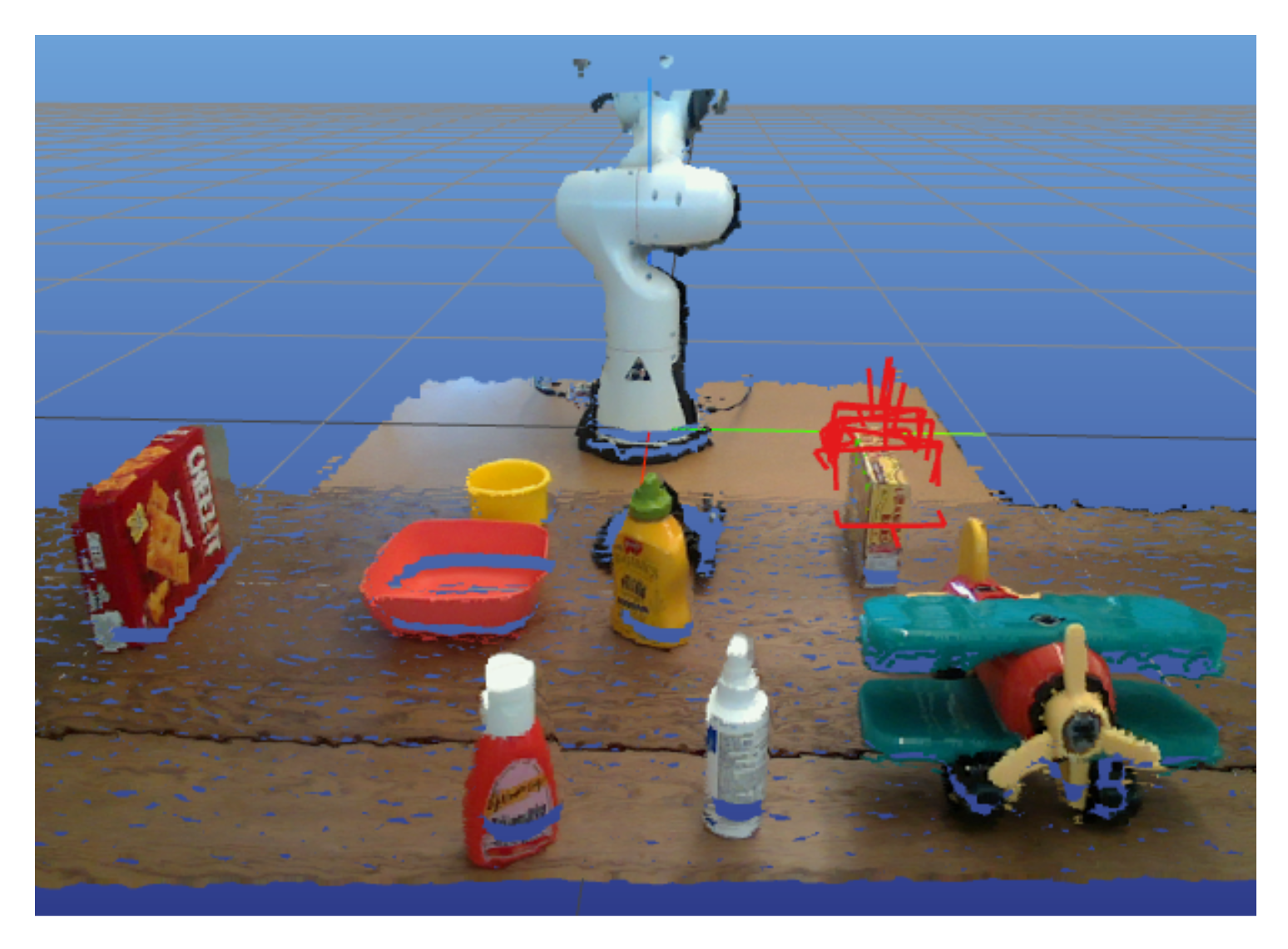} & 
    \includegraphics{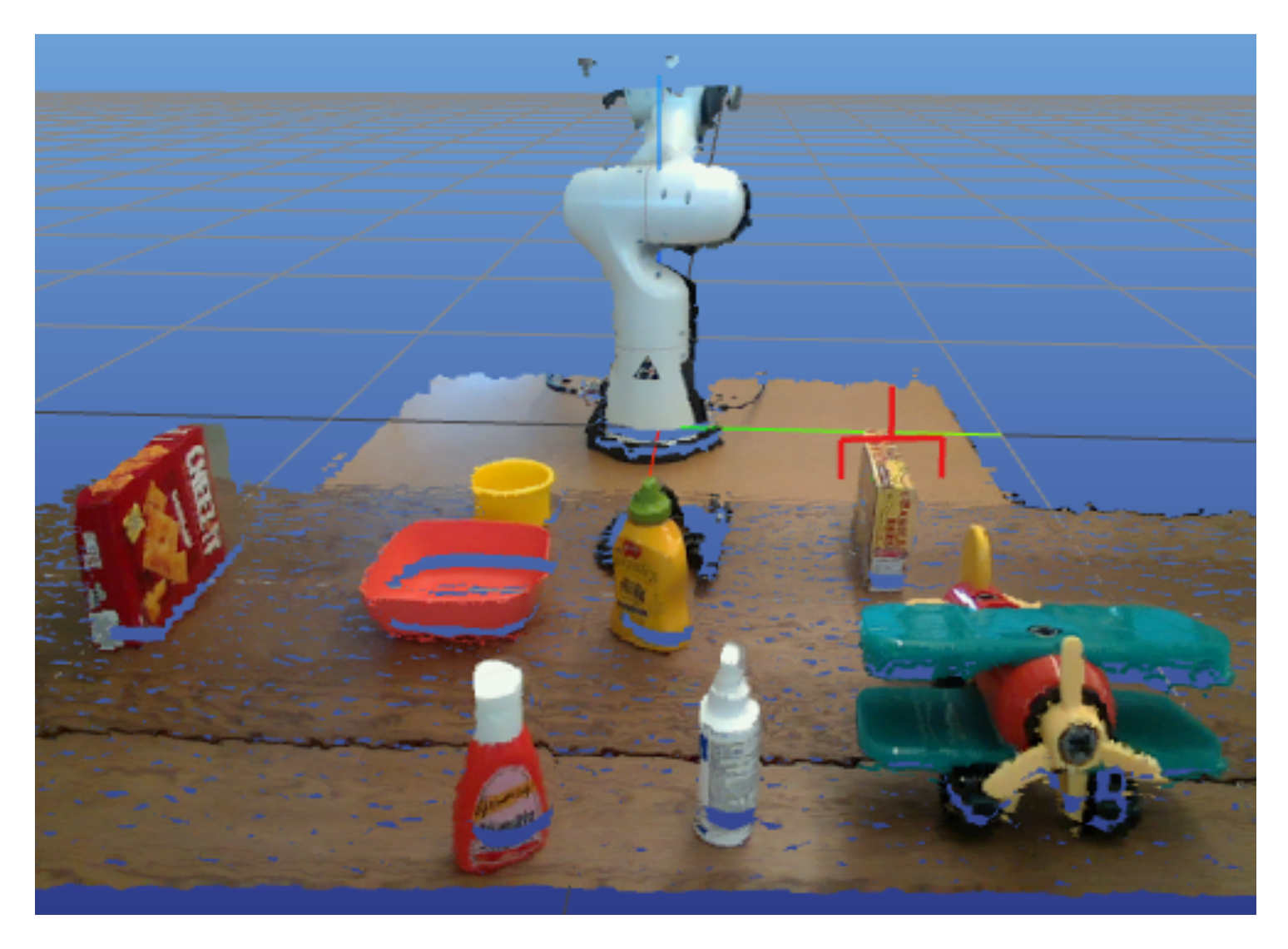} \\
  \end{tabular}
  \caption{Conversion from a 4 DoF to a 6 DoF grasp pose: (top-left) initial top-down grasp rectangle, (top-right) sampled contact points reprojected to 2D, (bottom-left) 6 DoF grasp proposals from Contact-GraspNet~\cite{sundermeyer2021contact}, (bottom-right) selected 6 DoF grasp.}
  \label{fig:grasping-representations}
\end{figure}

From an RGB-D image and a referring expression identifying a target object to grasp, language-driven grasping models must predict a stable grasp pose for a parallel gripper or dexterous hand. This poses three primary challenges: (i) identifying the target from textual cues (color, shape, spatial relations); (ii) selecting a collision-free grasp executable in real settings; and (iii) generalizing to unseen objects. Vision-language driven grasping is studied with both 4 DoF top-down and 6 DoF oriented grasps. While 6 DoF grasps are more accurate, annotating such grasp poses at a large scale to create real-world datasets is challenging. A number of works thus rely on image-based datasets with 4 DoF grasp rectangles. We design our method such that it can be used with both 4 DoF and 6 DoF grasping datasets. \cref{fig:grasping-representations} illustrates the two grasp formats and their transformation via an external proposal sampler. 


Formally, MapleGrasp predicts three dense maps, each of size \( H \times W \): a grasp quality map \( Q \in \mathbb{R}^{H \times W} \), a grasp width map \( w \in \mathbb{R}^{H \times W} \), and a grasp angle map \( \theta \in \mathbb{R}^{H \times W} \). The quality map assigns a score to each pixel, indicating the likelihood of a successful grasp. The width map encodes the optimal gripper width at each pixel, while the angle map represents the in-plane grasp orientation \( \theta \in [-\pi/2, \pi/2] \). A 2D rectangular grasp \( g_{\text{2D}} = (x^*, y^*, w^*, \theta^*) \) is extracted by selecting the pixel \( (x^*, y^*) = \arg\max_{(i,j)} Q_{i,j} \), and reading \( w^* = w_{x^*, y^*} \), \( \theta^* = \theta_{x^*, y^*} \). This rectangular grasp is projected onto the point cloud \( \mathcal{P} \) to identify a subset of points \( \mathcal{P}_g \subset \mathcal{P} \) within the grasp rectangle. A set of candidate 6-DoF grasp poses \( \{ G_k \} \) is generated using an independent grasp sampler. Each candidate pose \( G_k \) is evaluated by computing its intersection with \( \mathcal{P}_g \), and the final grasp \( G^* \) is chosen as the one that overlaps the maximum number of graspable points: \( G^* = \arg\max_k \, |\mathcal{P}_g \cap \mathcal{V}(G_k)| \), where \( \mathcal{V}(G_k) \) denotes volume swept by the gripper pose \( G_k \).

\subsection{Motivation for Mask Pooling}

Most LDRG methods predict an object mask and grasp pose simultaneously. A binary object segmentation mask already pinpoints the pixels referenced by the query, implicitly limiting where a valid grasp can lie. If the mask is correct, the grasp network need not evaluate locations outside it, shrinking search space and reducing errors. We treat the mask as an explicit \textit{guiding mechanism}: features are pooled within the segmented region before grasp prediction. 



\subsection{MapleGrasp}
\label{subsec:mpp}

Given an RGB image \( I \in \mathbb{R}^{H \times W \times 3} \) and a referring instruction \( T \), we compute vision-language embeddings using a pre-trained CLIP model: \( x_{\text{img}} = \text{CLIP}_{\text{img}}(I) \), \( x_{\text{text}} = \text{CLIP}_{\text{text}}(T) \). The visual embedding \( x_{\text{img}} \) is passed through a multi-modal Feature Pyramid Network \( \mathcal{F} \), and fused with the text embedding via cross-attention, resulting in \( x_{\text{fused}} = \text{CrossAttn}(\mathcal{F}(x_{\text{img}}, x_{\text{text}}), x_{\text{text}}) \). These embeddings are then processed through a series of convolutional and upsampling layers to produce a spatial feature map \( f \in \mathbb{R}^{H' \times W' \times C} \), where \( H' \) and \( W' \) are intermediate spatial resolutions and \( C \) is the channel dimension. We introduce \textit{Mask-Guided Feature Pooling} within a two-stage training framework to guide grasp prediction. In \textbf{Stage I (Segmentation Training)}, we freeze the grasp prediction heads and train the segmentation branch independently. The text embedding \( x_{\text{text}} \) is projected via a linear layer to obtain \( z \in \mathbb{R}^C \), which is then used to compute the object-centric mask via a per-pixel dot product: \( M_{i,j} = \sigma(f_{i,j}^\top z) \), yielding a soft binary mask \( M \in [0,1]^{H' \times W'} \). This stage allows mild overfitting to ensure accurate object localization for downstream pooling. In \textbf{Stage II (Mask-Guided Grasp Prediction)}, we train the full model end-to-end. The predicted mask \( M \) is used to modulate the visual features via element-wise multiplication, producing \( f_{\text{pooled}} = f \odot M \). These pooled features are refined and passed through three parallel projection heads to predict the grasp quality \( Q = \mathcal{H}_q(f_{\text{pooled}}) \), angle \( \theta = \mathcal{H}_\theta(f_{\text{pooled}}) \), and width \( w = \mathcal{H}_w(f_{\text{pooled}}) \).

Our goal was to support both 4-DoF and 6-DoF grasping with minimal architectural changes for future dataset compatibility. For image-only datasets with 4-DoF grasp pose annotations, we utilize the grasp rectangle to compute ground truth quality, angle, and width maps, following prior work~\cite{ocidvlg}. For RGB-D datasets with 6-DoF grasp poses, we first identify the 3D area on the target object where ground truth poses occur. This region is then projected into 2D to generate a rectangle over the graspable area of the object, which is subsequently treated in a manner analogous to the 4-DoF case. MapleGrasp computes the graspable region on the object, which is then overlapped with candidate 6-DoF grasp poses generated by an off-the-shelf grasp sampler, Contact-GraspNet~\cite{sundermeyer2021contact}. The candidate with maximum overlap is selected as the collision-free grasp for the target.

\begin{table*}[!h]
\centering
\small
\setlength{\tabcolsep}{2pt}  
\begin{tabular}{cccccccc}
\toprule
\textbf{Dataset} & \textbf{Real/Simulated} & \textbf{No. of Images} & \textbf{Ref.\ Text} & 
\textbf{Unique Obj.} & \textbf{Grasp Poses} & \textbf{Obj.\ Masks} & \textbf{Avg.\ Obj./Img.} \\
\midrule
Jacquard~\cite{jacquard}        & \cmark & 54K   & \xmark & --  & 967K   & \xmark & 1      \\
VMRD~\cite{8967869}            & \cmark & 4.5K  & \xmark & 31  & 51.5K  & \xmark & 3.5   \\
OV-Grasp~\cite{10802654}        & Both   & --    & 63K    & 117 & \xmark & \cmark & --     \\
ACRONYM~\cite{acronym}         & \xmark & --    & \xmark & 262 & 17.7M  & \xmark & --     \\
RoboRefIt~\cite{vlgrasp}       & \cmark & 10.7K & 50.7K  & 66  & \xmark & \cmark & 7.1     \\
OCID-VLG~\cite{ocidvlg}        & \cmark & 1.7K  & 89.6K  & 31  & 521K   & \cmark & 17.08  \\
GraspAnything++~\cite{graspanything++} & \xmark & 994K  & 10M    & 236 & 33M    & \xmark & 3.4    \\
\midrule
\textbf{Ours: RefGraspNet} & \cmark & 97K   & \textbf{12.25M} & 88  & \textbf{219M}   & \cmark & 9.57   \\
\bottomrule
\end{tabular}
\caption{Comparison of existing LDRG datasets. RefGraspNet provides 200M+ high quality grasps, object masks for each scene, and thus can be used to finetune both referring segmentation and grasping models.}
\label{tab:dataset_comparison}
\end{table*}

\subsection{Loss Function}
Prior work utilizes the Smooth L1 loss for pixel-wise prediction in visual grounding, defined per pixel \((\hat{x}_p, x_p)\) as:
\begin{equation}
L_{\mathrm{smoothL1}}(\hat{x}_p, x_p) = 
\begin{cases}
\tfrac{(\hat{x}_p - x_p)^2}{2\beta}, & \!\text{if }|\hat{x}_p - x_p|<\beta,\\
|\hat{x}_p - x_p| - \tfrac{\beta}{2}, & \!\text{otherwise.}
\end{cases}
\label{eq:smoothl1}
\end{equation}

Here, \( \hat{x}_p \in \mathbb{R} \) denotes predicted value at pixel \( p \), \( x_p \in \mathbb{R} \) is the ground-truth value, and \( \beta > 0 \) is a constant that determines the transition point between L2 and L1 loss regions.

We extend this to formulate weighted Smooth L1 loss \( w_p\,L_{\mathrm{smoothL1}}(\hat{x}_p, x_p) \), by assigning per‐pixel weights
\( w_p = 1 + \alpha\, Q_{\mathrm{gt},p} \), where \( Q_{\mathrm{gt},p} \in [0,1] \) is the ground‐truth \emph{quality map} indicating the object grasp region. By emphasizing suitable grasp regions, we enhance precise grasp estimation. Our experiments demonstrate that this modified loss function is beneficial in low-resource settings, where the dataset contains only single grasp annotation per object, ensuring more reliable and efficient learning.

\subsection{Metrics}
\label{subsec:metrics}


Language-driven grasp poses are evaluated based on their correctness, feasibility, and overall quality, given a language instruction and the corresponding scene image. Quantitative evaluation is performed using grasp success rate for both Top-1 and Top-K predictions. For 4-DoF grasping, a prediction is considered successful if its rotation angle differs from the ground-truth by less than \(30^\circ\), and its intersection-over-union (IoU) with the ground-truth grasp rectangle exceeds 0.25~\cite{graspmamba,maskgrasp}. For 6-DoF grasping, success is determined if the selected grasp proposal corresponds to the ground-truth grasp for the referred object~\cite{deshpande2025graspmolmo}.

%% file: sec/4_refgraspnetdataset.tex


\section{RefGraspNet Dataset: Learning to Grasp from 200M+ Grasp Poses}
\label{sec:refgraspnet}

To advance language-driven robotic grasping, we introduce RefGraspNet, a richly annotated dataset of over \textbf{219 million} high-quality 6 DoF grasp poses, derived from real-world cluttered scenes. Building on GraspNet-1B~\cite{fang2020graspnet}, which provides over one billion sampled grasps with force-closure scoring comprising 88 distinct household and industrial objects arranged in hundreds of densely cluttered table-top scenarios. Each scene captured from multiple calibrated viewpoints yielded object masks and 6 DoF grasp poses.  

\begin{figure}[!ht]
  \centering
   \includegraphics[width=\linewidth]{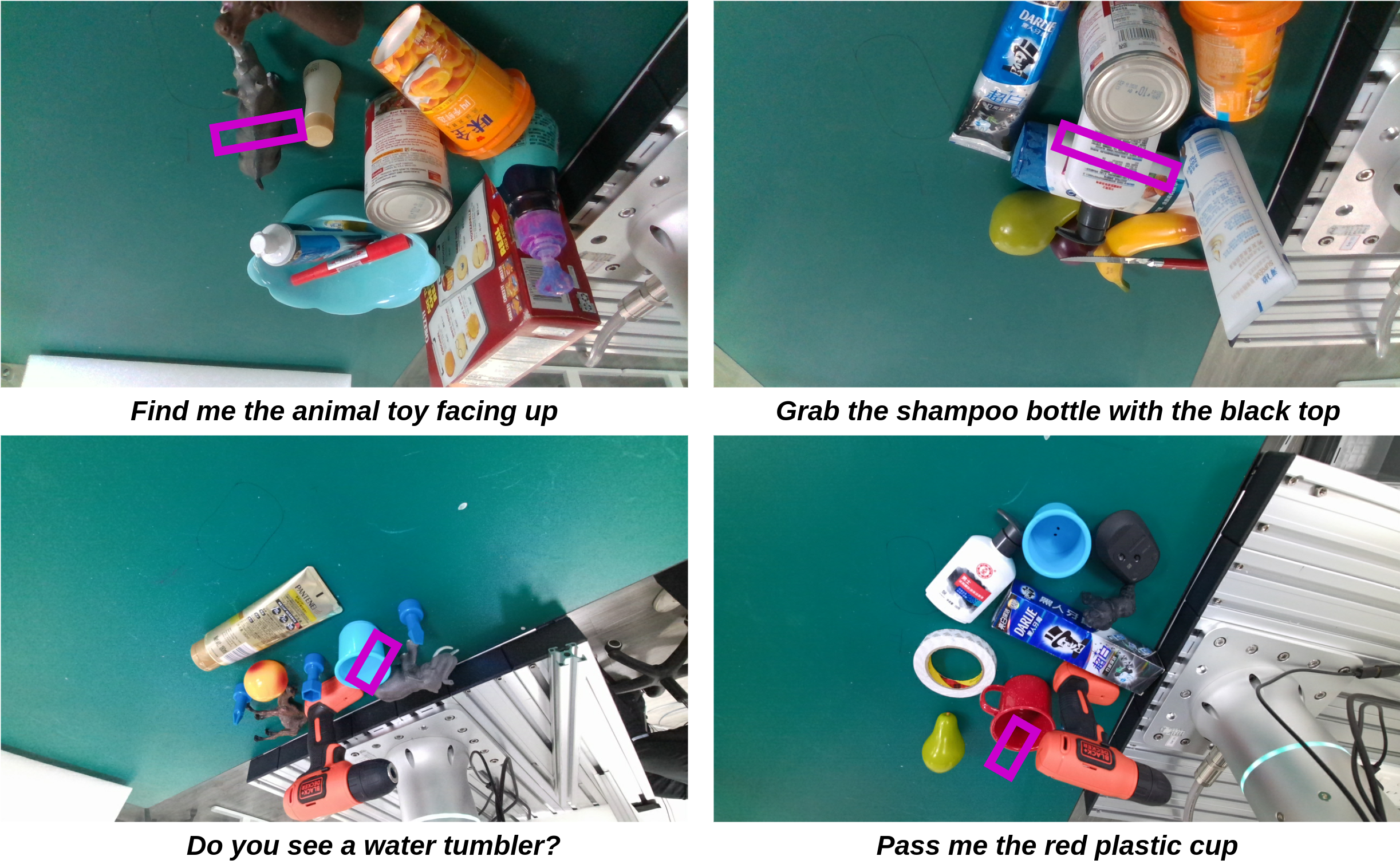}
   \caption{Samples from RefGraspNet showcasing cluttered scenes with multiple viewpoints, with diverse objects, distractors, and grasping instructions. Only top-down grasps shown for brevity.}
   \label{fig:refraspnet_examples}
\end{figure}

We automate generating referring expressions using DeepSeek-VL~\cite{lu2024deepseekvl}, an open-source VLM. Starting from ten instruction templates (e.g., “Grasp the \{object\},” “Locate the \{object\}”), we expanded object mentions with color, shape, and spatial qualifiers in multi-object layouts (e.g., “Grasp the red cup behind the yellow bowl”). This yielded over 12.25M unique referring phrases, ensuring robust grounding of grasp commands in cluttered environments.  Grasp candidates were filtered using a 70\% force-closure confidence threshold, which was selected based on extensive manual validation, resulting in 219 million validated 6 DoF grasps—roughly eight times larger than the previous largest language-driven grasp dataset. We provide comprehensive comparisons in \cref{tab:dataset_comparison}, demonstrating improvements in scale, scene complexity, and linguistic variety. To support empirical studies, we split our objects into 70\% “seen” (62 objects) and 30\% “unseen” (26 objects), and generate train/val, test-seen, and test-unseen splits. This design enables evaluation of generalization to novel objects and arrangements. Examples are shown in \cref{fig:refraspnet_examples}. 


%% file: sec/5_experiments.tex
\begin{table*}[t]
\centering
\begin{tabular}{c c c cc}
\toprule
\textbf{Model} & \multicolumn{1}{c}{\textbf{Epochs to Conv.}} & \multicolumn{1}{c}{\textbf{OCID-VLG}} & \multicolumn{2}{c}{\textbf{RefGraspNet}} \\
& & & \textbf{Test Seen} & \textbf{Test Unseen} \\
\midrule
DetSeg + CLIP~\cite{detseg} & 22 & 28.12 / 39.21 & 40.19 / 41.56 & 27.18 / 27.20 \\
GR-ConvNet + CLIP~\cite{gr-convnet} & 15 & 9.73 / 15.41 & 34.19 / 46.56 & 30.19 / 32.17 \\
SSG + CLIP~\cite{xu2023instance} & 41& 33.51 / 34.70 & 48.78 / 50.10 & 14.36 / 14.89 \\
Molmo+SAM2+GraspNet & --- & 63.22 / 69.14 & 68.85 / 70.27 & 66.23 / 69.41 \\
CROG~\cite{ocidvlg} & 50 & 77.22 / 87.71 & 85.32 / 86.49 & 70.81 / 71.99 \\
HiFi-CS~\cite{bhat2024hificsopenvocabularyvisual} & 44 & 70.54 / 79.12 & 73.27 / 74.55 & 62.13 / 63.28 \\
ETRG~\cite{yu2024parameter} & 49 & 82.28 / 91.12 & 86.13/88.12 & 68.11 / 70.05 \\
\midrule
\textbf{Ours: MapleGrasp} & 32 & \textbf{88.15 / 92.98} & \textbf{89.86 / 91.67} & \textbf{76.92 / 77.67} \\
\bottomrule
\end{tabular}
\caption{Benchmarking MapleGrasp across OCID-VLG and RefGraspNet datasets for 4-DoF and 6-DoF vision-language driven grasping respectively. Scores reported as grasping accuracy for top-1/top-5 predictions (see \cref{subsec:metrics}). }
\label{tab:combined_benchmark}
\end{table*}

\vspace{-8pt}
\section{Experiments}
\label{sec:experiments}

We evaluate MapleGrasp on OCID-VLG and RefGraspNet, while comparing it to prior LDRG and VLA methods in both simulation and real-world settings.

\subsection{Benchmark Comparisons}
\label{subsec:benchmark_comparisons}
The OCID-VLG corpus~\cite{ocidvlg} contains 1763 highly cluttered indoor tabletop RGB-D scenes with 31 unique objects. Objects have annotated 4-DoF grasp poses and segmentation masks, associated with referring expressions describing attributes such as color, shape, and relative position. With 89K unique (RGB-Text-Mask) tuples, it is the largest publicly available dataset for LDRG\footnote{GraspAnything++ is open-sourced; however, its public subset is much smaller than the full training data, making fair benchmarking impossible.}. We also provide a benchmark on our new contributed RefGraspNet corpus.


We evaluate various CLIP-based architectures on both datasets: CROG~\cite{ocidvlg} utilizes CLIP embeddings with cross-attention, followed by MLP layers, to simultaneously predict object masks and grasp maps. HiFi-CS~\cite{bhat2024hificsopenvocabularyvisual} employs a frozen CLIP model with FiLM-based fusion and upsampling to generate object masks, followed by grasp estimation using a separate frozen network. ETRG~\cite{yu2024parameter} introduces trainable adapters within the CLIP model to adapt it for grasping, and is trained on additional data to segment referring expressions and predict referring affordance. All baselines are trained for 50 epochs to ensure a fair comparison (see \cref{tab:combined_benchmark}). We also include a strong zero-shot baseline that combines Molmo~\cite{deitke2024molmopixmoopenweights} for identifying the referred object, SAM 2~\cite{ravi2025sam} for generating the object mask, and GraspNet~\cite{fang2020graspnet} for predicting both 4-DoF and 6-DoF grasp poses.  MapleGrasp achieves a 7\% relative improvement in Top-1 success rate over previous state-of-the-art methods, demonstrating higher precision in grasp prediction in OCID-VLG. In RefGraspNet, MapleGrasp outperforms baselines for both seen and unseen objects, with a larger performance margin observed in the unseen setting. We attribute this to our closely coupled dual-stage training, which is able to provide partial object masks in unseen objects, which is then used by the grasp detection module to generate graspable object regions and accurate grasp poses. MapleGrasp trains more efficiently, converging in 32 epochs (\(\sim\)8 hours), compared to 45–50 epochs (12–16 hours) required by other baselines on an RTX 4090 GPU.




\begin{table*}[ht]
\centering
\small
\setlength{\tabcolsep}{4pt} 
\begin{tabular}{>{\centering\arraybackslash}m{2.5cm}
                >{\centering\arraybackslash}m{4cm}
                >{\centering\arraybackslash}m{4cm}
                >{\centering\arraybackslash}m{4cm}}
\toprule[1pt]
\textbf{Referring Text} & \textbf{Molmo+SAM+GraspNet} & \textbf{CROG} & \textbf{MapleGrasp} \\
\midrule[0.5pt]
Grasp the rightmost red food box & 
\includegraphics[width=\linewidth]{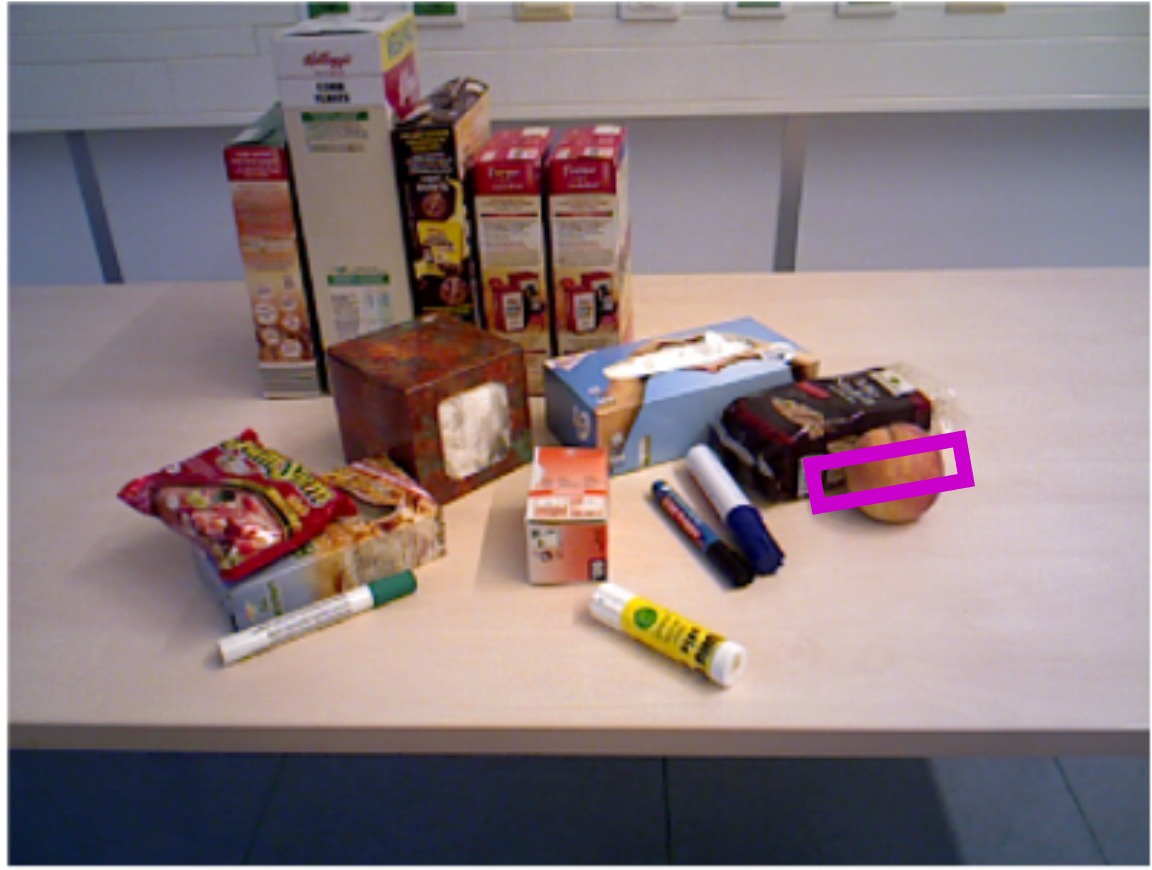} &
\includegraphics[width=\linewidth]{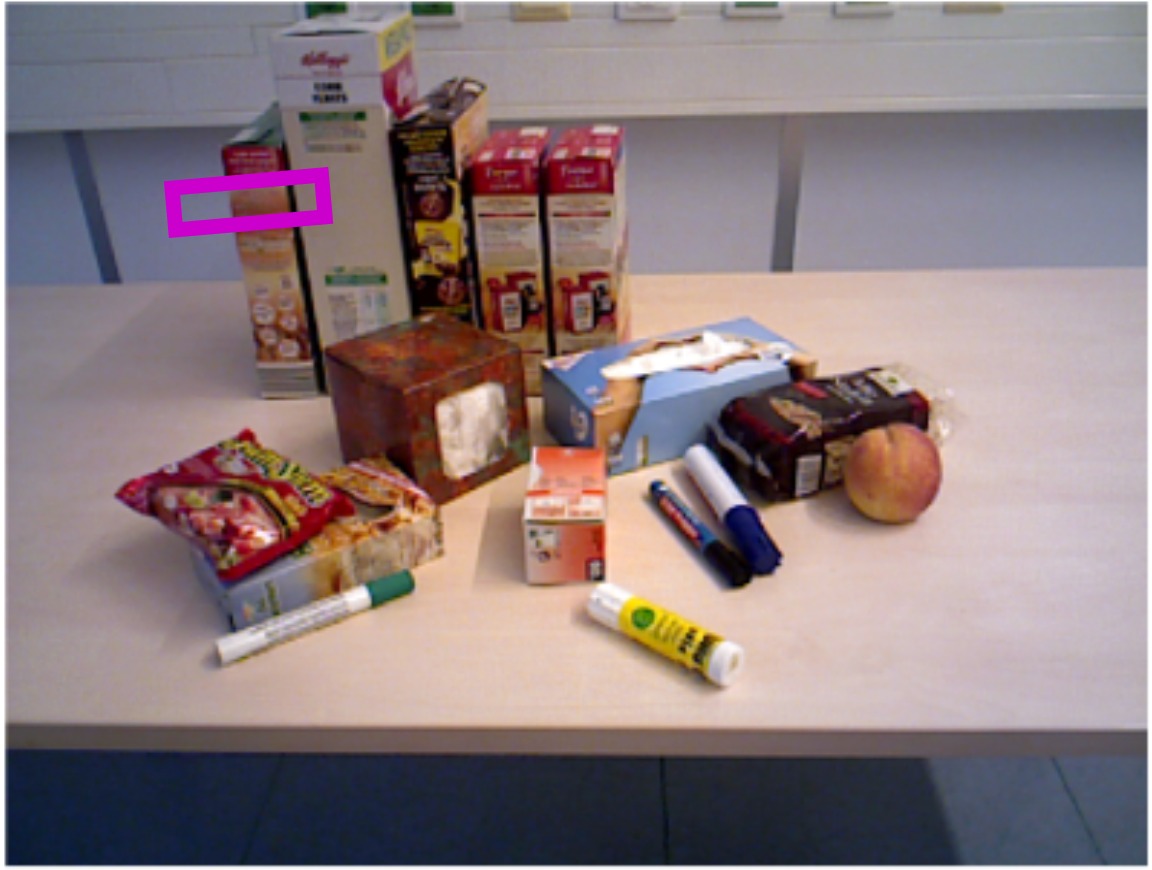} &
\includegraphics[width=\linewidth]{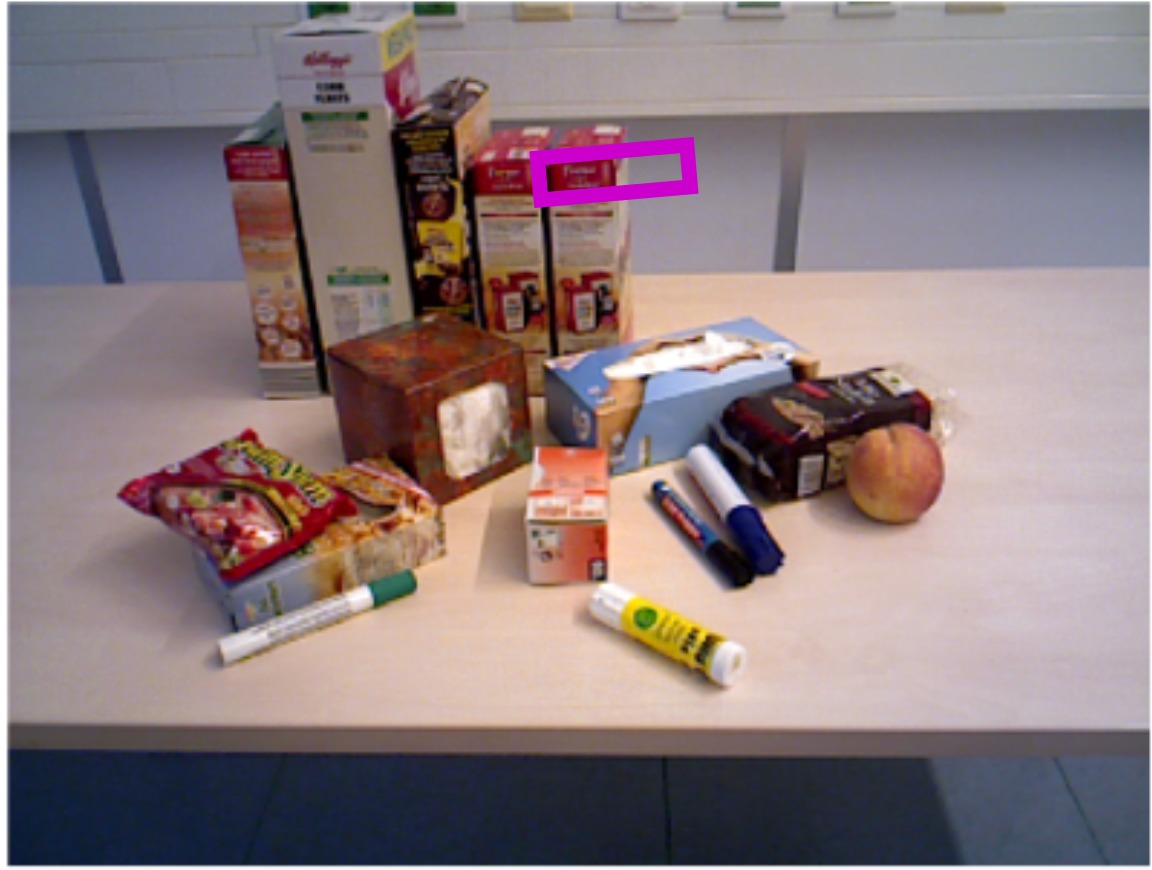} \\
\midrule

Grab the transparent food bag &
\includegraphics[width=\linewidth]{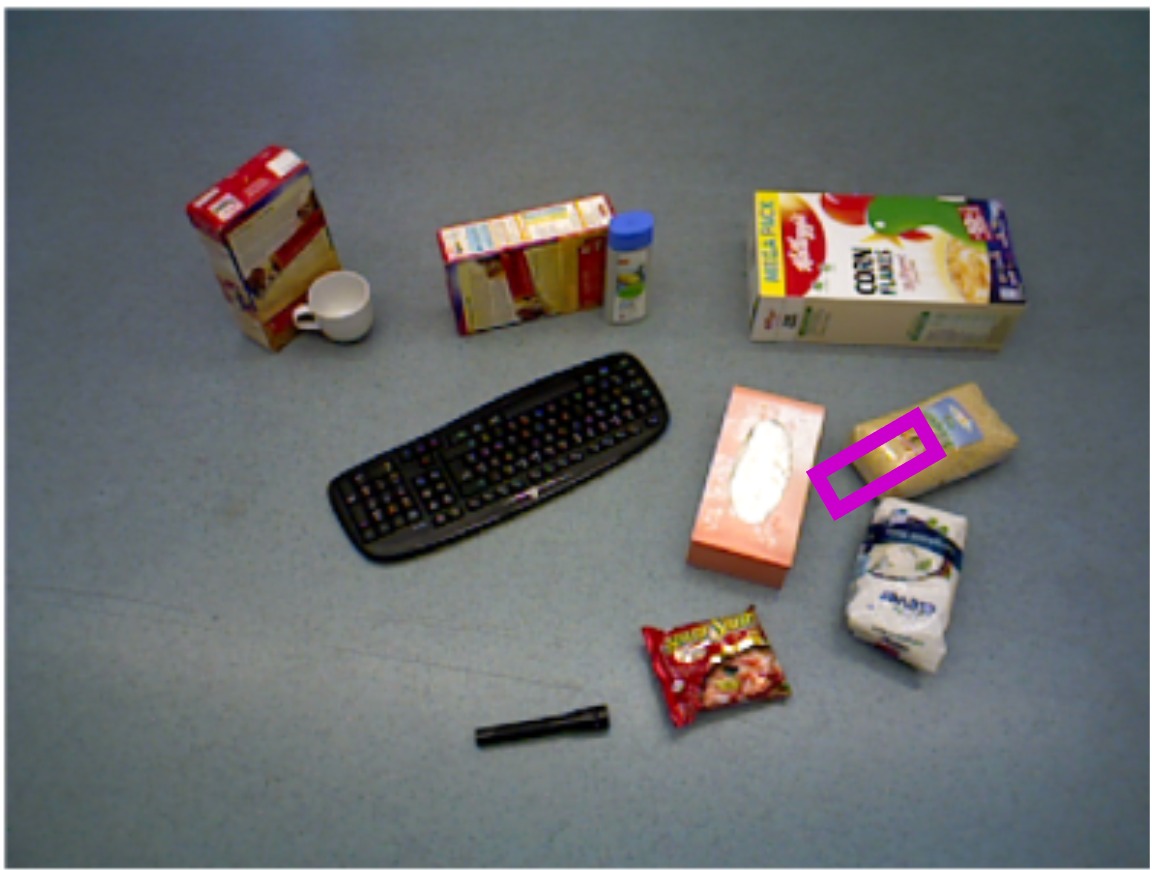} &
\includegraphics[width=\linewidth]{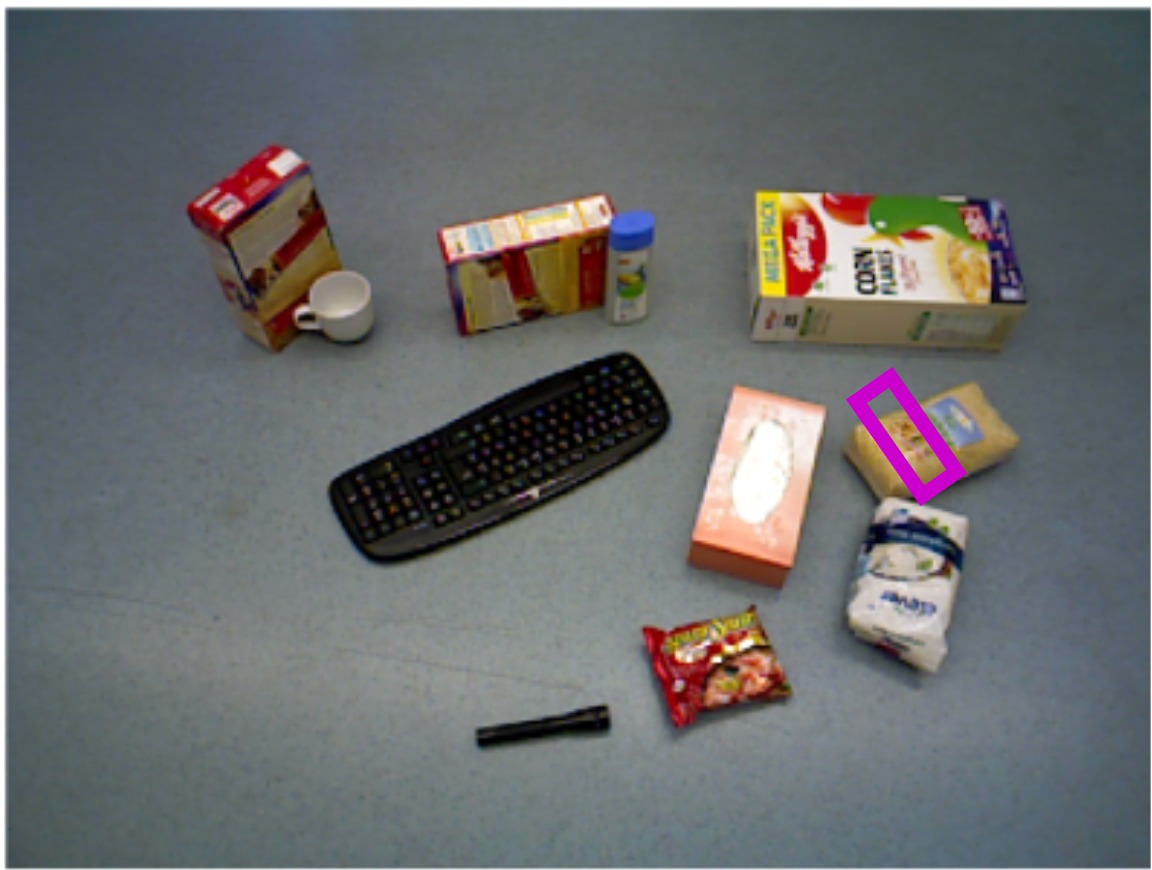} &
\includegraphics[width=\linewidth]{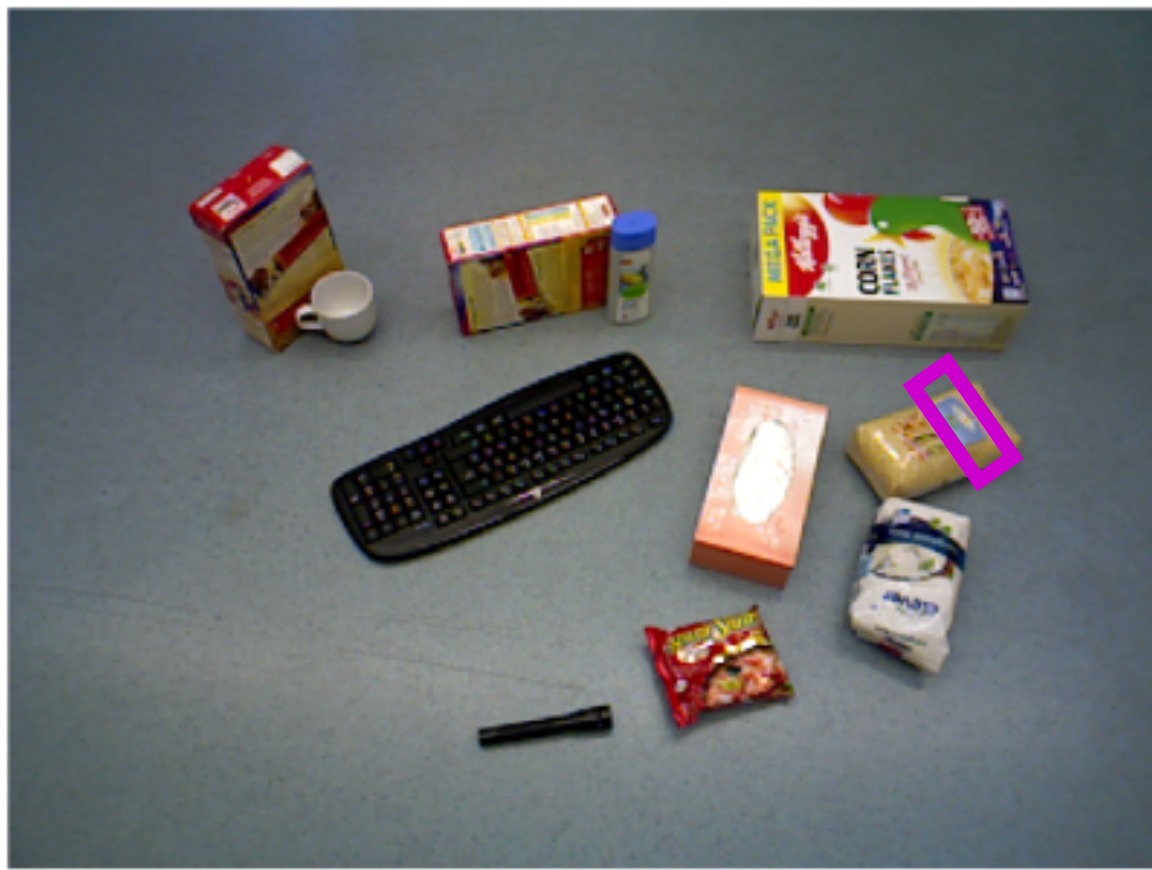} \\
\midrule

Find the banana furthest away from me &
\includegraphics[width=\linewidth]{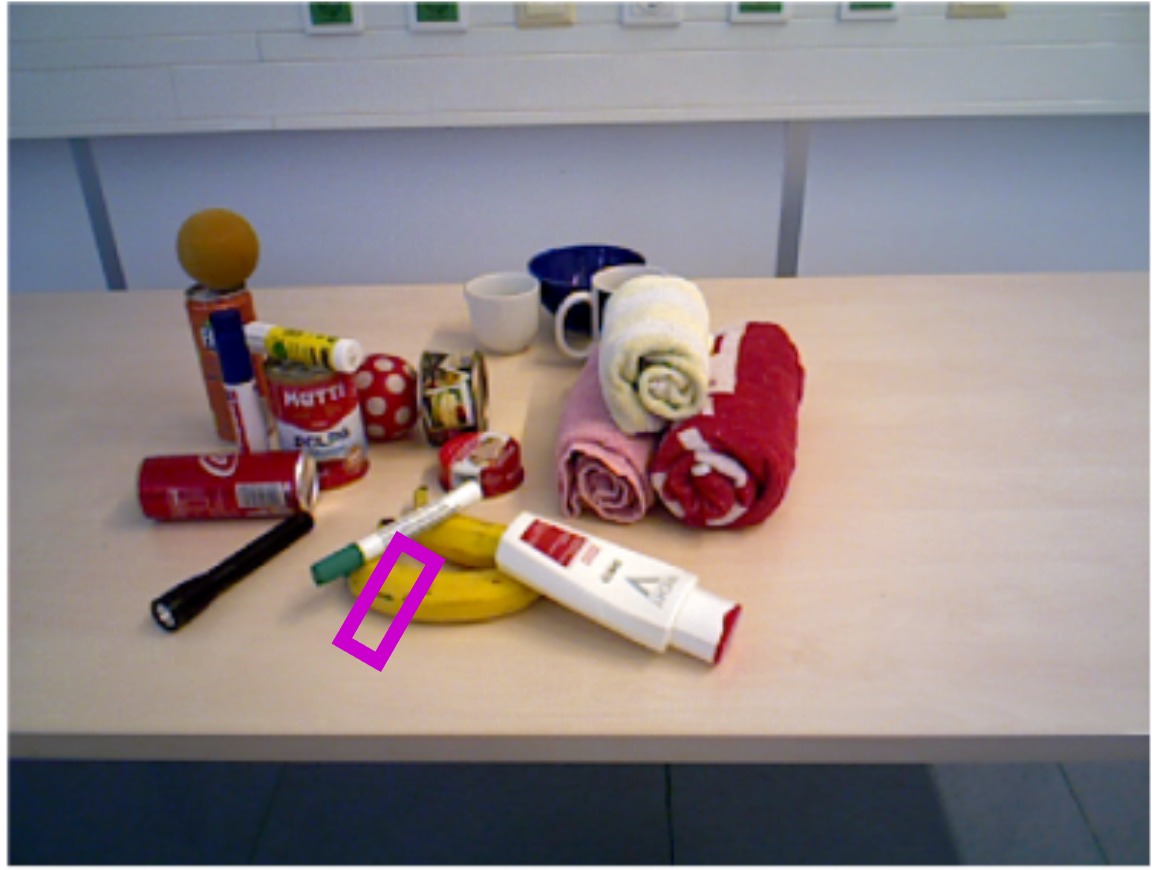} &
\includegraphics[width=\linewidth]{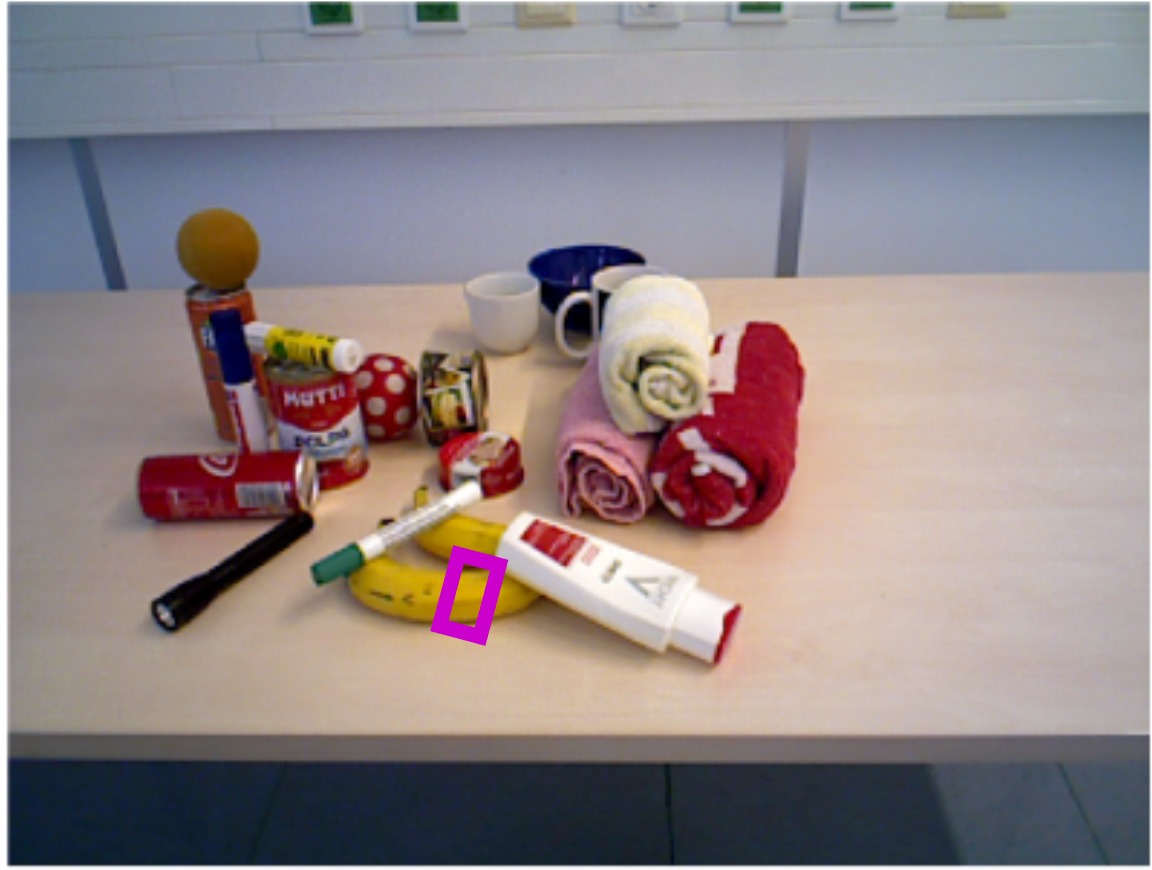} &
\includegraphics[width=\linewidth]{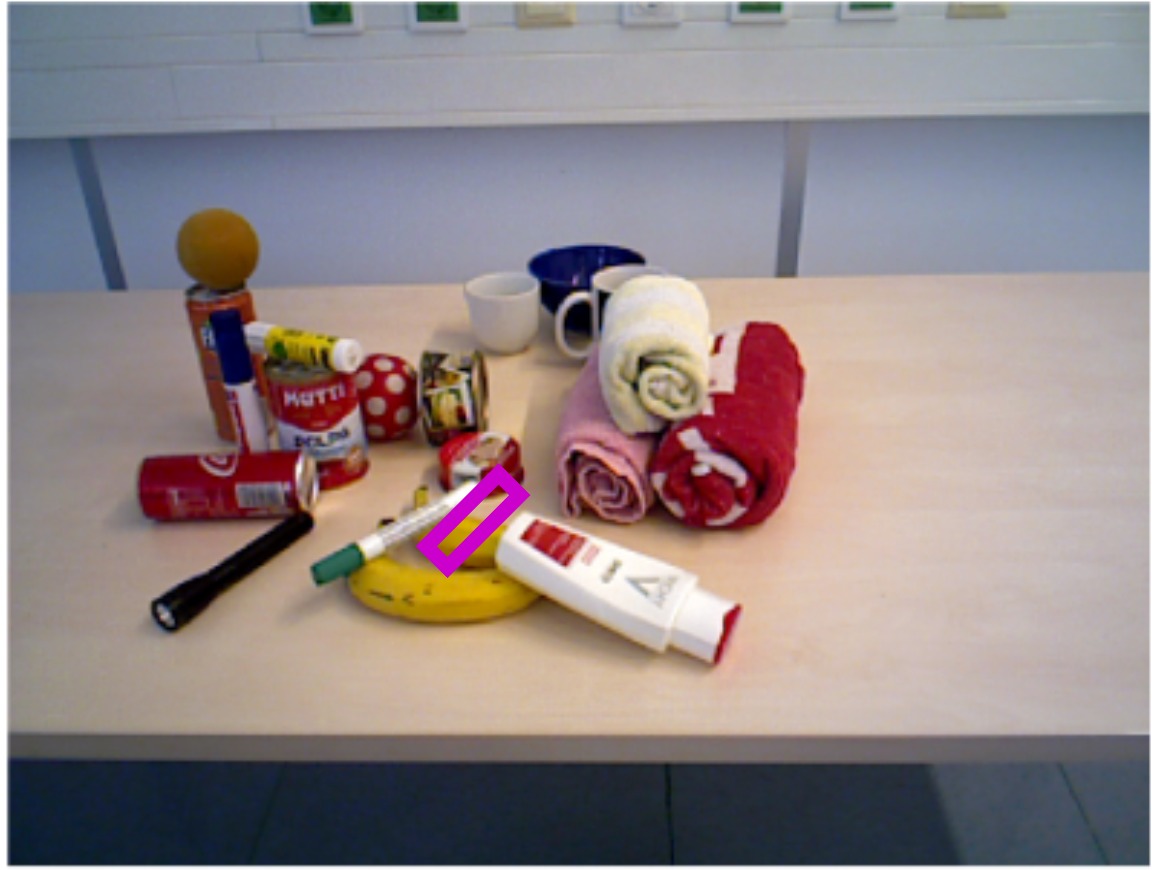} \\
\midrule

Pass the tissues to the rear left of the blue and black marker &
\includegraphics[width=\linewidth]{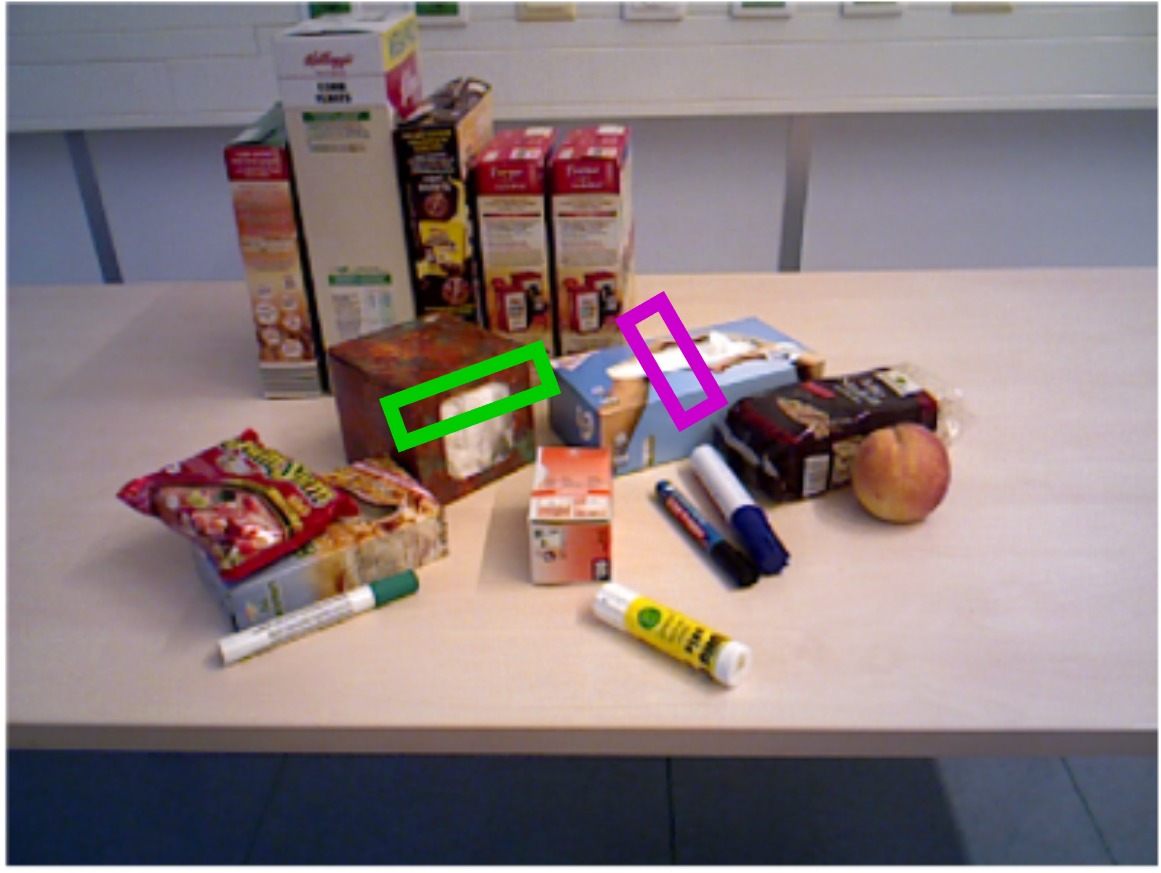} &
\includegraphics[width=\linewidth]{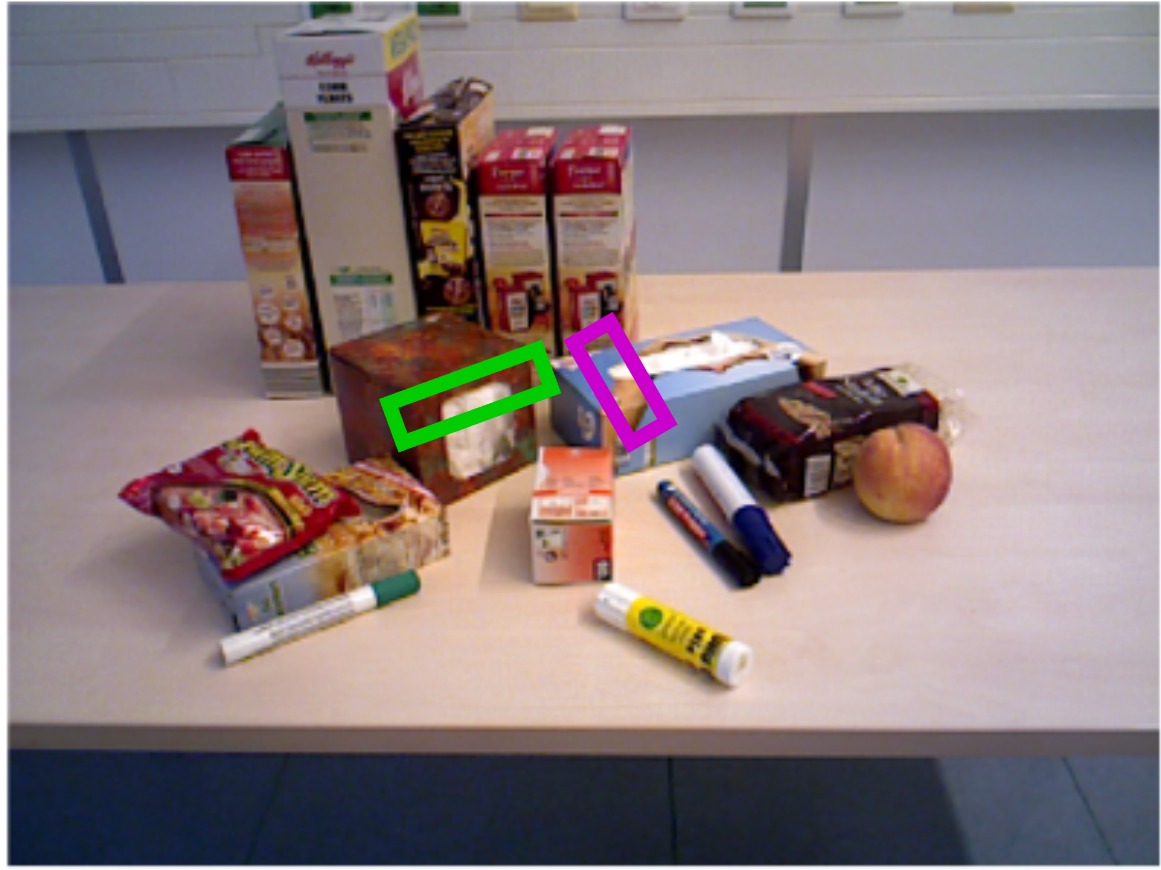} &
\includegraphics[width=\linewidth]{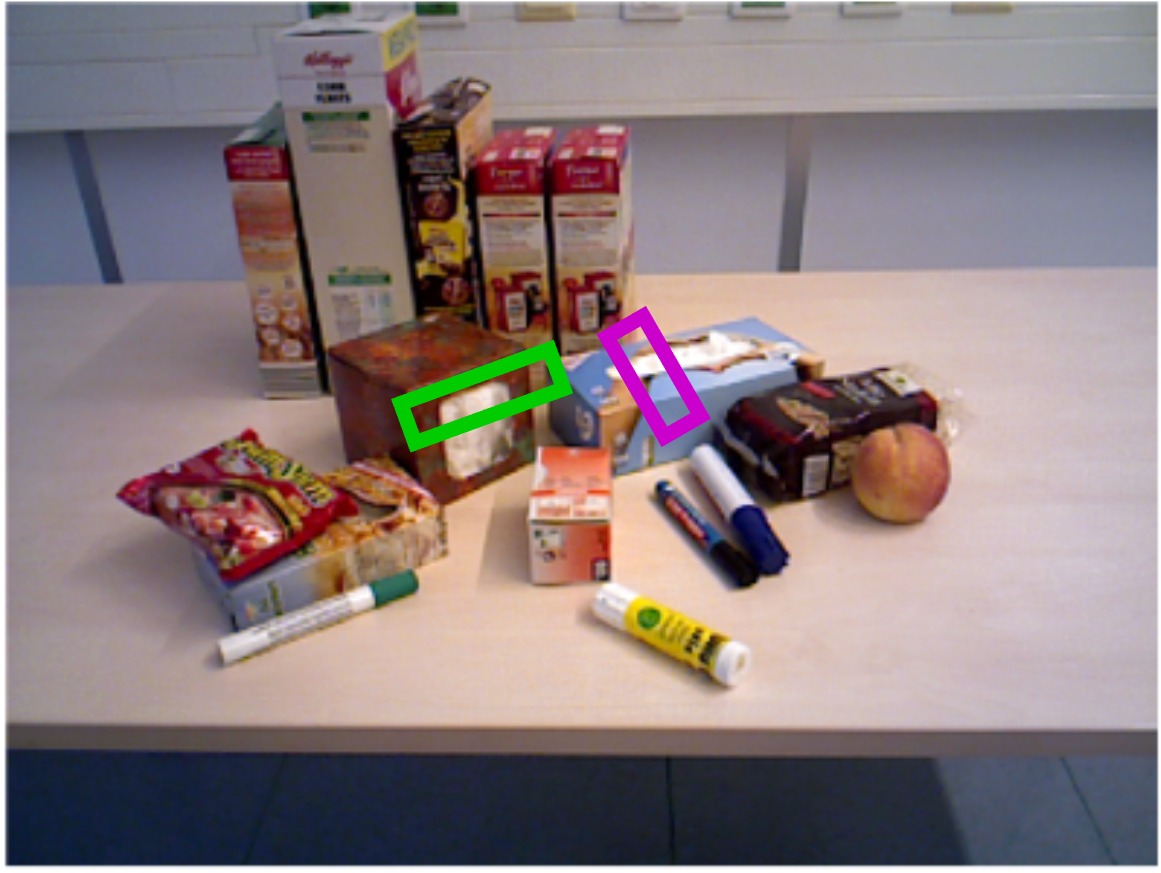} \\
\bottomrule[1pt]
\end{tabular}
\caption{Qualitative comparisons of LDRG approaches under complex referring queries. The zero-shot baseline (first column) struggles to identify referred object due to distractors. MapleGrasp (last column) can predict an accurate and accessible grasp for queried objects in the first three rows. Last row shows an example where all models fail to recognize the correct object (ground truth in green).}
\label{tab:comparison_examples}
\end{table*}


\subsection{Qualitative and ablation studies}
\label{subsec:ocidvlg-ablations}

We provide comparisons in (\cref{tab:comparison_examples}).  In row 1 and 3, we observe that the off-the-shelf approach fails to accurately predict a grasp on the referred object. This occurs because the query contains attributes like ``rightmost" and ``furthest" increasing the complexity of referring. Since the masks are predicted incorrectly, the downstream GraspNet predicts an incorrect grasp. While CROG recognizes these objects better, having seen a variety of referring attributes during training, often fails to localize the grasp within the target object. MapleGrasp’s mask-guided predictions are accurate in first three rows, but fail for occluded objects and ambiguous queries, where the mask alone cannot localize the grasp pose (row 4). More examples in supplementary Sec.~$\mathcal{S}$2.

\begin{figure}[!ht]
  \centering
   \includegraphics[width=
   \linewidth]{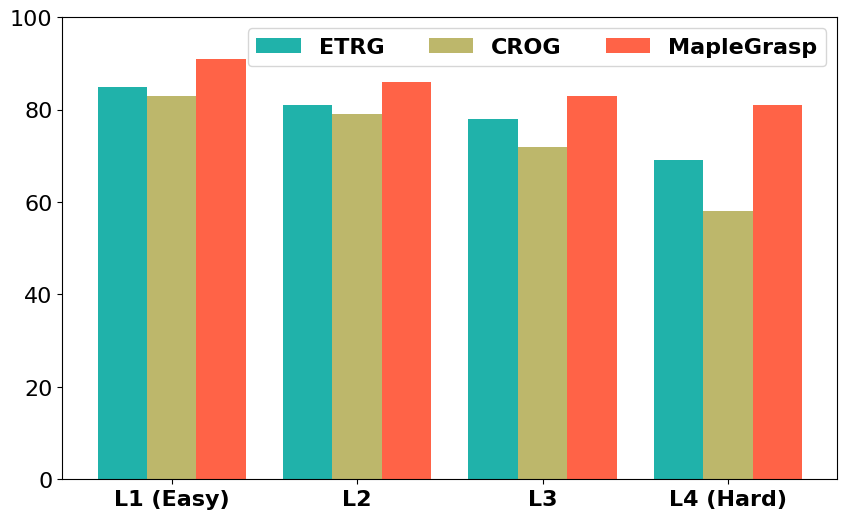}
   \caption{Comparisons across text inputs: from level 1 with just object names to level 4 with color, shape, position attributes}
   \label{fig:referring-attrib-analysis}
\end{figure}


To compare methods across language complexity, we group test queries by the number of object attributes they mention. For example, ``Grasp the \underline{red} \underline{circular} box near the \underline{top right} of the image" contains three attributes that an LDRG model must exploit to locate a grasp. Using a popular named entity recognition model, GLiNER~\cite{zaratiana-etal-2024-gliner}, we count attributes in the RefGraspNet test split and assign queries to four bands—L1 EASY (one attribute) through L4 HARD (four). \cref{fig:referring-attrib-analysis} plots Top-1 accuracy for MapleGrasp and two baselines: accuracy declines with added attributes, yet MapleGrasp consistently remains ahead.

We conduct ablation studies on MapleGrasp (\cref{tab:ablations}) to analyze its effects due to architectural modifications. Results highlight the importance of accurate RES for mask pooling; using Molmo, which achieves lower IoU (50-60\%) on object mask detection, negatively impacts grasp accuracy due to complex referring queries requiring task-specific visual grounding. While using frozen CLIP improves object identification, it reduces grasp scores due to insufficient grounding of VL features for grasp detection. Replacing cross-attention with MLP-Mixers~\cite{NEURIPS2021_cba0a4ee}, does not improve performance. Furthermore, replacing the weighted smooth L1 loss with a standard smooth L1 loss leads to reduced scores due to slower convergence. All ablations use a fixed number of training epochs.

\begin{table}[!ht]
\centering
\setlength{\tabcolsep}{4pt} 
\resizebox{\columnwidth}{!}{%
\begin{tabular}{lcc}
\toprule
\textbf{Method} & \textbf{Top-1} & \textbf{Top-5} \\
\midrule
MapleGrasp & \textbf{88.15} & \textbf{92.90} \\
\midrule
(1) Molmo+SAM for Mask Pooling & 78.69 & 81.27 \\
(2) Frozen CLIP Layers          & 73.14 & 75.57 \\
(3) No Cross-Attn, MLP-Mixer    & 85.42 & 91.54 \\
(4) Standard Smooth L1 Loss     & 84.35 & 90.23 \\
\bottomrule

\end{tabular}}
\caption{Ablation study on the OCID-VLG corpus.}
  \label{tab:ablations}
\end{table}

\begin{figure}[!ht]
  \centering
   \includegraphics[width=\linewidth]{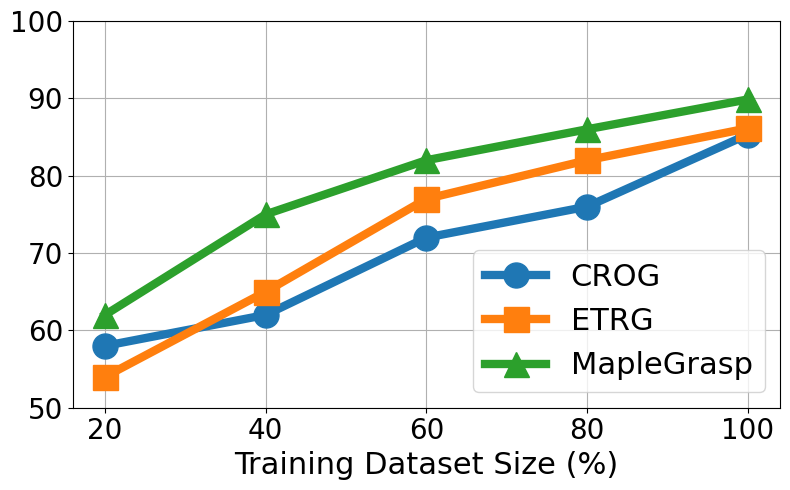}
   \caption{Training data efficiency comparisons across baselines on RefGraspNet. Scores reported as grasping success rate in \%.}
   \label{fig:training-efficiency-refgraspnet}
\end{figure}

Lastly, we evaluate the training efficiency of MapleGrasp in comparison to two baseline methods. In this experiment, we vary the size of the training dataset and assess model performance after training on 20\%, 40\%, 60\%, 80\%, and 100\% of the RefGraspNet corpus. The results, presented in \cref{fig:training-efficiency-refgraspnet}, demonstrate that MapleGrasp consistently outperforms the baselines on the test set as the dataset size increases, with particularly notable improvements observed after training on just 20--40\% of the data. These findings suggest that incorporating referring object masks for pooling vision-language feature maps in grasp detection enhances both training effectiveness and sample efficiency.

\vspace{-5pt}
\subsection{Cross-Dataset Testing}
\label{subsec:cross-dataset-testing}

To assess generalizability of models trained on RefGraspNet, we cross-test them on the OCID-VLG benchmark. We compare CROG and MapleGrasp \cref{tab:cross_dataset} and our findings reveal that models trained on RefGraspNet generalize better than those trained solely on OCID-VLG. 



\begin{table}[t]
\centering
\small
\setlength{\tabcolsep}{3.0pt} 
\begin{tabular}{l c cc cc}
\toprule
& & \multicolumn{2}{c}{\textbf{OCID-VLG}} & \multicolumn{2}{c}{\textbf{RefGraspNet}} \\
\textbf{Model} & \textbf{Train-Dataset} & \textbf{Top-1} & \textbf{Top-5} & \textbf{Top-1} & \textbf{Top-5} \\
\midrule
CROG & OCID-VLG & 77.2 & 87.7 & 41.8 & 42.9 \\
     & RefGraspNet & \textbf{68.2} & \textbf{73.8} & 85.32 & 86.49 \\
MapleGrasp & OCID-VLG & 88.2 & 92.9 & 43.2 & 45.6 \\
           & RefGraspNet & \textbf{79.5} & \textbf{81.7} & 89.2 & 89.7 \\
\bottomrule
\end{tabular}
\caption{Cross-dataset evaluation results (highlighted). Models trained on RefGraspNet generalize better than OCID-VLG.}
\label{tab:cross_dataset}
\end{table}

\vspace{-5pt}
\subsection{LIBERO Physics Simulation Experiments}
\label{subsec:libero-results}

We test MapleGrasp in the LIBERO simulation environment in a table-top setting. The grasp poses generated by MapleGrasp are executed with an inverse kinematics velocity controller described in supplementary Sec.~$\mathcal{S}$3. We compare our modular grasping+trajectory optimization method with end-to-end VLAs. While VLAs are trained on tele-operated demonstrations of successful tasks, we use these observations to identify stable grasp poses for referred objects. We only benchmark the object grasping skill. For example, consider the task ``Put the wine bottle on the rack." We consider the task to be successful if the robot grasps the ``wine bottle" and lift it above the ground. Success rate is measured manually and averaged with 10 trial runs per task.

\begin{table}[!h]
\centering
\small
\setlength{\tabcolsep}{2pt} 
\begin{tabular}{c c c c}
\toprule
\textbf{Model} & \textbf{SPATIAL} & \textbf{GOAL} & \textbf{OBJECT} \\
\midrule
Diffusion Policy & 77 & 69 & \textbf{92} \\
Octo & 78 & 84 & 88 \\
OpenVLA& 84 & 79 & 90\\
MapleGrasp & \textbf{87} & \textbf{85} & 90\\
\bottomrule
\end{tabular}
\caption{Libero simulation results across SPATIAL, GOAL and OBJECT task suites. Scores are reported using success rate}
\label{tab:libero-results}
\end{table}


Experiments are conducted on three task suites from the LIBERO benchmark: SPATIAL, GOAL, and OBJECT. For more details about the tasks, refer to supplementary Sec. ~$\mathcal{S}$4. Results in \cref{tab:libero-results} compare our method with VLA and diffusion baselines. Our results demonstrate that our grasping+control pipeline achieves similar accuracy to recent VLAs in grasping tasks. Interestingly, decoupling grasping and control improves generalization, as trajectory-based imitation learning methods fail to complete unseen tasks. However, our model successfully determines grasp poses for similar but unseen objects, allowing the control algorithm to generate a stable trajectory without requiring video demonstrations. Cross-dataset testing in \cref{tab:libero-crossdataset} reveals the capabilities of our approach in handling unseen queries across task suites. Since our LDRG model is only concerned with predicting grasp poses and we leverage a controller to determine the trajectory, our method can be deployed without learning from tele-operated demonstrations.

\begin{table}[t]
\centering
\small
\setlength{\tabcolsep}{3pt} 
\begin{tabular}{c c c c}
\toprule
\textbf{Model} & \textbf{Train-Dataset} & \textbf{GOAL} & \textbf{OBJECT} \\
\midrule
OpenVLA & GOAL & 79 & 0\\
  & OBJECT & 12 & 90\\
MapleGrasp & GOAL & 85 & \textbf{62}\\
  & OBJECT & \textbf{68} & 90\\
\bottomrule
\end{tabular}
\caption{Cross-dataset results show OpenVLA struggles on seen objects in new tasks, while MapleGrasp generalizes better.}
\label{tab:libero-crossdataset}
\end{table}


\subsection{Real-Robot Experiments}
\label{subsec:real-robot-experiments}
\vspace{-10pt}
\begin{figure}[h]
  \centering
   \includegraphics[width=\linewidth]{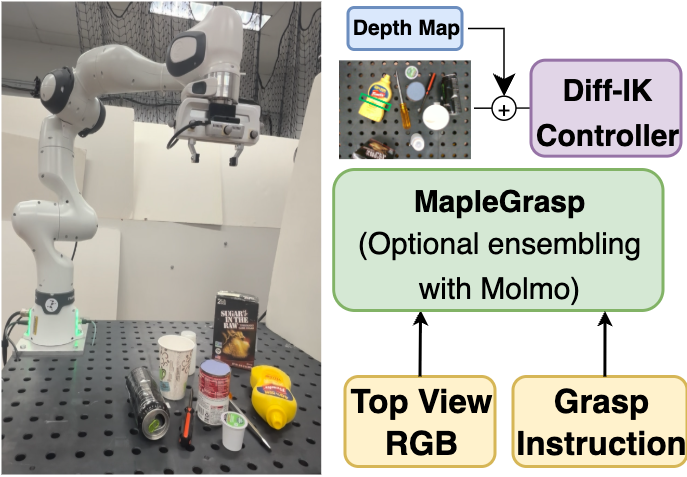}
   \caption{Setup for robot trials using 7 DoF franka arm.}
   \label{fig:real-robot-experiments}
\end{figure}

We evaluate various LDRG methods in a real-world robotic setup using a Franka arm with a camera mounted at the gripper, as shown in \cref{fig:real-robot-experiments}. All models were fine-tuned on a combined dataset of RefGraspNet and OCID-VLG. For evaluation, we collected 15 objects, 10 seen and 5 unseen to create randomly arranged tabletop scenes (see supplementary Sec. ~$\mathcal{S}$5). To test capabilities of all baselines, we performed separate experiments trying to grasp seen and unseen objects in the presence of distractors (objects that look similar to the target object either in shape, color, or type). Each grasping instruction was tested 10 times and averaged. This resulted in N = 300 (15x2x10) robot trials. Results are presented in \cref{tab:real-robot-experiments}. We consider a trial to be successful if the robot grasps object and lifts it 30cm above the surface.

In the first column (seen objects) all methods tackle the simplest case, having seen these items during training. MapleGrasp leads with a 93$\%$ success rate, ahead of CROG and HiFi-CS. The second column adds clutter; varied language queries and distractors lower accuracy for every model, yet MapleGrasp still tops the chart at 87$\%$. The final two rows probe zero-shot grasping of unseen items. Here the Molmo+SAM2+GraspNet baseline rivals MapleGrasp because Molmo—a VLM pre-trained on millions of images spanning thousands of categories—can robustly identify novel targets for GraspNet. To harness this open-world knowledge we form MapleGrasp+Molmo: at inference we compare MapleGrasp’s segmentation mask with Molmo’s out-of-vocabulary prediction, and when MapleGrasp’s confidence is low we keep the mask with the highest overlap before pooling features for the grasp decoder. This strategy lifts success on cluttered scenes with unseen objects to 73$\%$, the best overall. Our two-stage training uses object masks as priors, enabling ensembling with an open-set VLM (see supplementary video for examples).


\vspace{-5pt}
\begin{table}[h]
\centering
\small  
\setlength{\tabcolsep}{2.0pt}  
\begin{tabular}{l cc cc}
\toprule
& \multicolumn{2}{c}{\textbf{Seen Objects}} & \multicolumn{2}{c}{\textbf{Unseen objects}} \\
\textbf{Model} & \textbf{Single} & \textbf{Cluttered} & \textbf{Single} & \textbf{Cluttered}\\
\midrule
Molmo+SAM2+GraspNet & 70 & 61 & 71 & 66\\
HiFi-CS & 72 & 63 & 65 & 59 \\
CROG & 85 & 73 & 69 & 54\\
\midrule
MapleGrasp               & \textbf{93} & \textbf{87} & 72 & 65\\
MapleGrasp+Molmo & 91 & 85 & \textbf{74} & \textbf{73}\\
\bottomrule
\end{tabular}
\caption{Real robot experiments. Success rate measured across 10 trials for every grasp instruction.}
\label{tab:real-robot-experiments}
\end{table}

%% file: sec/6_conclusion.tex
\vspace{-8pt}
\section{Conclusion}
\label{sec:conclusion}
Our results highlight mask-guided feature pooling as a simple yet effective technique for language-driven robotic grasping. By selectively pooling predictions within target regions, MapleGrasp reduces misidentifications and grasp errors in cluttered environments. Our two-stage training strategy, which first predicts segmentation masks and subsequently refines grasp predictions, further enhances accuracy and computational efficiency. Compared to end-to-end vision-language-action models, our modular approach demonstrates superior scalability and robustness, and enables rapid adaptation using readily available vision-language models.  Moreover, these systems eliminate the need for time-intensive teleoperated demonstrations and can be customized using a small corpus of RGB-text-grasp annotations. Real-world validation confirms significant improvements in grasp accuracy and efficiency.
